\documentclass{article}

\usepackage{lmodern}

\usepackage[utf8]{inputenc} 
\usepackage[T1]{fontenc}    
\usepackage{hyperref}       
\usepackage{url}            
\usepackage{booktabs}       
\usepackage{amsfonts}       
\usepackage{nicefrac}       
\usepackage{microtype}      
\usepackage{xcolor}         
\usepackage{graphicx}

\usepackage{amssymb}
\usepackage{makecell}

\usepackage{titling}
\predate{}
\postdate{}

\title{AI Model Utilization Measurements For Finding Class Encoding Patterns}

\author{Peter Bajcsy\thanks{point of contact} \and
Antonio Cardone \and
Chenyi Ling \and
Philippe Dessauw \and
Michael Majurski \and 
Tim Blattner \and
Derek Juba \and
Walid Keyrouz \and \\
National Institute of Standards and Technology, Gaithersburg, MD 20899 \\
\texttt\{peter.bajcsy, antonio.cardone, chenyi.ling, philippe.dessauw, michael.majurski, \\
 timothy.blattner, derek.juba, walid.keyrouz\}@nist.gov \\
}
\date{} 

\begin{document}

\maketitle

\begin{abstract}
  
This work addresses the problems of (a) designing utilization measurements of trained artificial intelligence (AI) models and (b) explaining how training data are encoded in AI models based on those measurements. The problems are motivated by the lack of explainability of AI models in security and safety critical applications, such as the use of AI models for classification of traffic signs in self-driving cars. We approach the problems by introducing theoretical underpinnings of AI model utilization measurement and understanding patterns in utilization-based class encodings of traffic signs at the level of computation graphs (AI models), subgraphs, and graph nodes. Conceptually,  utilization is defined at each graph node (computation unit) of an AI model based on the number and distribution of unique outputs in the space of all possible outputs (tensor-states). In this work, utilization measurements are extracted from AI models, 
which include \emph{poisoned} and \emph{clean} AI models. In contrast to clean AI models, the poisoned AI models were trained with traffic sign images containing systematic, physically realizable, traffic sign modifications (i.e., \emph{triggers}) to change a correct class label to another label in a presence of such a \emph{trigger}. We analyze class encodings of such clean and poisoned AI models, and conclude with  implications for trojan injection and detection. 

\end{abstract}

\section{Introduction}
\label{introduction}

The \emph{motivation} of this work lies in the lack of interpretability and explainability of artificial intelligence (AI) models in security and safety critical applications.  
For instance, regular traffic signs and any physically realizable trigger modifications represent intended and hidden encoded classes in a classification AI model (e.g., a yellow sticky on top of a \emph{STOP} traffic sign as a trigger changing the label from the intended \emph{STOP} to the target \emph{65 m/h} traffic sign classes \cite{Xu2019}). 
Our lack of understanding of how classes are encoded in AI models for classifying traffic signs poses a safety threat in self-driving cars because  AI models can contain such injected triggers causing misclassification.  In addition to the application-specific motivation, methods for explainable AI are motivated in general by regulatory agencies, end users, decision makers, and engineers as they provide utilities for bias detection, trust in predictions, suitability for deployment, debugging, and recourse \cite{Arrieta2020}, \cite{Lakkaraju2020}. 

We introduce the \emph{terminology} used in this paper early on due to a varying usage of published terms in a broad spectrum of theoretical contributions to AI.  We will refer to an AI model as a computation graph that (a) is a directed graph representing a math function and (b) consists of subgraphs. A subgraph is a subset of graph vertices (or graph nodes) connected with edges in the parent graph. Graph nodes of a computation graph are computation units (or graph components) that perform linear or non-linear operations on input data, (e.g., convolution, tangent hyperbolic activation function, and maximum value operation). In our work, the names of the AI models (or architectures) are adopted from literature since we are not creating any custom computation graphs.
The input and output data at each computation unit are multidimensional arrays denoted as tensors.  When an image from a class $c$ flows through a computation graph, each computation unit generates real-valued tensors called class activations (the term is derived from an activation function that decides whether a neuron should be activated or not). A tensor generated by input images has dimensions reflecting a number of images (batch size), channels, rows, and columns. For a batch size equal to one, a tensor can be interpreted as a hyperspectral image. 
In our work, a class activation mapping is thresholding, and binarized channel values in one tensor are denoted as a tensor-state with rows $\times$ columns of tensor-state values.

The \emph{objectives} of this work are 
(1) to define utilization-based class encodings and AI model fingerprints, 
(2) to measure class encodings in architectures beyond small models (e.g., LeNet model with 60K parameters) and toy datasets, such as MNIST (Modified NIST dataset with 70K images of size $28 \times 28$ pixels), and 
(3) to identify encoding patterns (motifs) that discriminate AI models without and with hidden classes (denoted as clean and poisoned AI models). 
Our \emph{ultimate objective}  is to identify and decompose AI model computation graphs into subgraphs (subnetworks) that serve specific detection, segmentation, classification, or recognition purposes. 
Building a library of subgraphs with semantic interpretations, such as a subgraph for wheel detection, creates opportunities to custom-design AI architectures for specific training datasets and avoid exploring a huge search space of graphs as done in the work of Ying et al. \cite{nasbench2019}.
This objective is aligned with the search for visual patterns in a collection of visualizations of AI layers and neurons under the OpenAI Microscope project \cite{openAImicroscope2022}.
By understanding class encoding patterns, one can additionally benefit from reduced AI model storage and inference computational requirements via more efficient network architecture search \cite{nasbench2019} with advanced hardware \cite{Justus2019}. Furthermore, one can improve expressiveness of AI model architectures via design \cite{Lu2017} and efficiency measurements \cite{schaub2020} or one can assist in diagnosing failure modes \cite{Bontempelli2021}.

This work addresses the \emph{problems} of (a) designing utlization measurements of trained AI models and (b) explaining how training data are encoded in AI models based on those measurements.
We approach the problems and address the three objectives as follows: 
 \begin{enumerate}
\item  Define a class encoding by introducing a utilization measurement at each computation unit in an AI model computation graph.
\item Form a class encoding as a vector of utilization measurements at all computation units of an AI model.
\item Search for class encoding patterns by training and analyzing clean and poisoned AI models together with their training datasets, as well as by visualizing their patterns in AI model computation graphs, subgraphs, and tensor-state spaces.  
\end{enumerate}

Conceptually, utilization of any computation unit is related to a ratio of the number of different outputs (tensor-state values) activated by all training data points over the maximum number of possible outputs by the computation unit. Such utilization-based class encodings are useful as statistical representations of complex trained AI models for (a) classifying a large number of AI models as clean or poisoned, and (b) reducing the search space for understanding class's unique and overlapping patterns. We use a set of tensor-states at each graph node and for each training image as a baseline representation of one trained AI model. With such a baseline representation, one can visually validate correctness of any conclusions derived from utilization-based class encodings for varying class characteristics, application-specific datasets, and AI model architectures. 

\begin{figure}
\includegraphics[
  width=12cm,
  height=8cm,
  keepaspectratio,
]{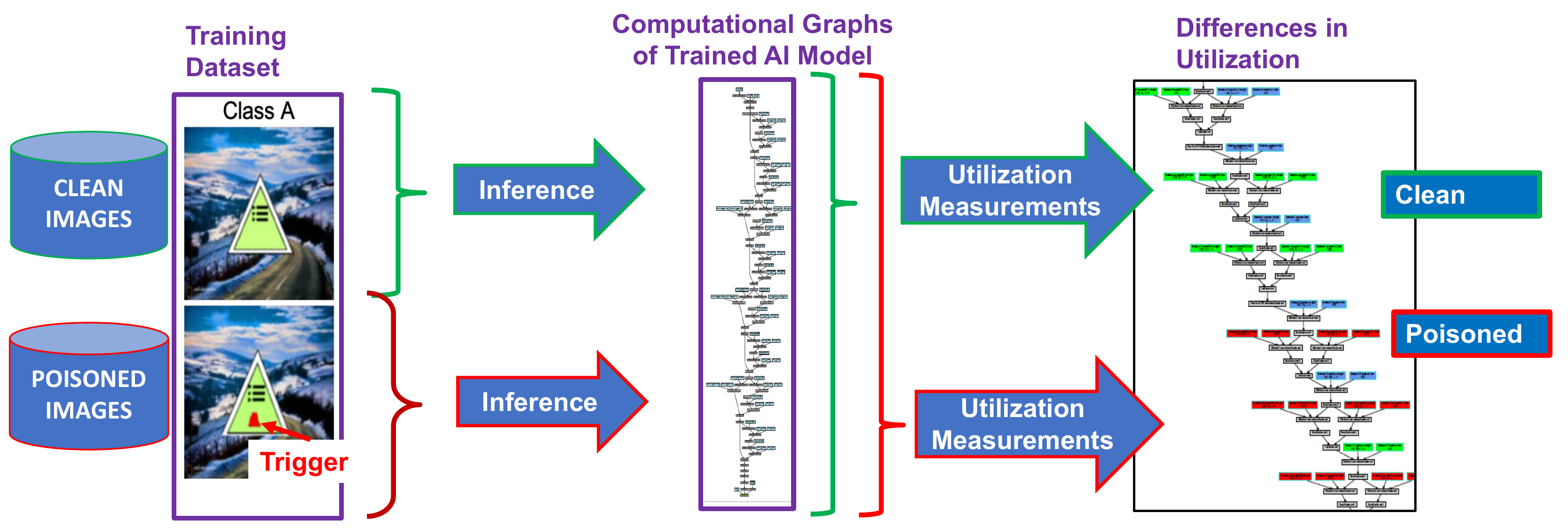}
  \centering
  \caption{A high-level workflow for identifying utilization patterns in an AI computation graph of ResNet18 architecture for clean and poisoned classes}
  \label{fig:01}
\end{figure}

Figure~\ref{fig:01} shows a high-level workflow for identifying discriminating patterns of class encodings in clean and poisoned AI models.
The left side in  Figure~\ref{fig:01} illustrates "Training Dataset" consisting of clean (Class A) and poisoned (Class B) training images with a small red polygon denoted as a trigger (or poison). The left side could also be replaced with two clean classes as we aim to identify patterns of class encodings unique to each class and common for any two classes.
Training images representing each class are inferenced. During the inference of images from the same class, a vector of utilizations over all graph computation units is recorded and denoted as a class encoding. Differences in class encodings can be visualized by color-coded AI computation graphs to contrast class encodings (e.g., clean and poisoned or clean Class A and Class B - see the right side of  Figure~\ref{fig:01}).



The key \emph{challenges} in addressing the problems lie in (1) integrating theoretical knowledge about neural networks to define utilization-based class encodings, (2) computing class encodings within an allocated time (e.g., 15 min per AI model), with limited computational resources and over hundreds of thousands of training images, and (3) visualizing and interpreting class encoding patterns for a variety of AI model architectures at the granularity of AI model graphs, subgraphs, and tensor-states.
All three challenges are intensified by increasing sizes of training datasets and by advancing complexities of AI architectures as enumerated with examples in Table~\ref{table:01}. As of today, AI architectures are
(a) very complex in terms of the number of parameters (from 60K parameters in LeNet model \cite{Khan2019}, to common networks having millions and billions of parameters, such as 160 billion reported by Trask et al. \cite{Trask2015}, and bleeding-edge networks with trillion-parameters in AI language models \cite{transformers2021}), 
(b) very heterogeneous in terms of types of computation units in computational AI graphs, and
(c) high dimensional in terms of data tensors generated by AI graph computation units.

\begin{table}
  \caption{Problems and their complexity challenges for AI models available from TrojAI Challenge, Rounds 1-4 \cite{IARPA2020}}
  \label{table:01}
\centering
\begin{tabular}{|c | c |}
\hline
Problems & Complexity Challenges  \\
\hline
\makecell{How to define  AI model utilization?} & \makecell{tensor-states in AI models \\ with $\approx 10^{12}$ parameters}  \\
\hline
\makecell{How to characterize class encoding of \\each class via utilization of \\AI model computation units?} &  \makecell{$\approx 10^5$ inferences\\per AI model} \\
\hline
\makecell{What AI model computation units\\are critical for\\class encodings?} &   \makecell{$\approx 10^3$ AI model\\fingerprints}           \\
\hline
\end{tabular}
\end{table}


The fundamental underlying \emph{assumptions} of our approach lie in the fact that tensor-state statistics at each graph node from all activations with training images per class can gain us insights into a mathematical function defining the mapping between class-defining training images and class labels. The tensor-state statistics are quantified based on capacity versus utilization metrics. Such metrics capture the concept of a successful defense against backdoor attacks by graph pruning as reported by Liu et al. \cite{Liu2018a} and are assumed to reveal a presence or absence of hidden classes (triggers or backdoor attacks). Although symbolic representations of subgraphs are still under investigation by Olah et al. \cite{olah2020zoom}, \cite{olah2017feature}, we also assume that the utilization-based characterization of subgraphs may have a relationship with symbolic descriptions of image parts (e.g., subgraphs encode a traffic sign shape) and hence presence or absence of trojans can be detected by finding patterns in utilization-based color-coded graphs and subgraphs.      

The main \emph{novelties} of this work are in the definition, measurement design, and pattern searching in  utilization-based clean and poisoned class encodings. The main \emph{contributions} are in utilization measurement placements for a variety of AI architectures, and in explainable clean and poisoned AI models at the granularity levels of AI model graphs, subgraphs and tensor-states.  
Our work leveraged interactive Trojan and AI efficiency simulations enabled by the Neural Network Calculator tool \cite{bajcsy2021} and web-accessible AI models generated for the TrojAI Challenge computer vision rounds \cite{IARPA2020}.  

The paper is organized as follows. Section~\ref{related_work} explains the relationship of this work to past efforts. Section~\ref{methods} presents the theoretical underpinnings of AI model utilization measurements, the design reasoning, and the methodology for finding class encoding patterns in trained AI models. While Section~\ref{experimental_results} documents experimental data, numerical results, and visualizations, Section~\ref{discussion} provides an interpretation of the results in the context of trojan detection and injection.
Finally, Section~\ref{summary} summarizes lessons learned and outlines future work.

\section{Related Work}
\label{related_work}

The problem of explainable AI is very broad and the term \emph{explainable} is still debated in philosophical texts \cite{Gilpin2018} (``What is an Explanation?''). A comprehensive survey of explainable AI has been published by Arrieta et al. \cite{Arrieta2020} and extensive teaching materials have been made available by Lakkaraju et al. \cite{Lakkaraju2020}. 
Our approach can be related to ``Explanation of Deep Network Representation'' (roles of layers, individual units, and representation vectors) according to the Deep Learning-specific taxonomy presented by Arrieta et al. in \cite{Arrieta2020}, Fig. 11.
Our utilization-based approach is inspired by exploring relationships between biological neural circuits and AI model computation graphs as discussed by Olah et al. in \cite{olah2020zoom}.  Next, the related work is presented with respect to the three formulated problems.

Our work on \emph{defining utilization} is related to the  past work on measuring neural network efficiency \cite{schaub2020}, \cite{bajcsy2021}, which is rooted in neuroscience and information theory. In the work of Schaub and Hotaling \cite{schaub2020}, neural efficiency and artificial intelligence quotient (aIQ) are used to balance neural network performance and neural network efficiency while inspired by the neuroscience studies relating efficiency of solving Tetris task and brain metabolism during the task execution \cite{Haier1992IntelligenceAC}. In the work of Bajcsy et al. \cite{bajcsy2021}, an online simulation framework is used to simulate efficiencies of small-size neural networks with a variety of features derived from two-dimensional (2D) dot pattern data.  In contrast to the previous work \cite{schaub2020}, \cite{bajcsy2021}, our theoretical framework defines and reasons about class encodings, AI model fingerprints, and metrics for finding class encoding patterns for much more complex AI models and training datasets. 

Following the categorization in the survey on interpreting inner structures of AI models \cite{Rauker2022survey}, the \emph{utilization measurements} can be related to concept vectors  whose goal is to associate directions in latent space with meaningful concepts. In the work of Fong and Vedaldi \cite{Fong2018}(Network 2 Vector) and Bau et al. \cite{Bau2017} (Network Dissection), the distribution of activation maps at each convolutional unit as inputs pass through is used to determine a threshold. Threshold-based segmented activation maps are compared across concepts. In contrast to the previous work \cite{Fong2018},\cite{Bau2017}, our utilization measurements are computed at all computation units in an AI model, the activation maps are binarized at zero, and statistics are computed over a distribution of tensor-states (including the binarized activation maps from convolutional units). Our approach does not use any inserted modules like in concept whitening \cite{Chen2020} to align the latent space with concepts. Furthermore, our approach does not project class activation maps to create saliency maps \cite{Selvaraju2019}, \cite{Adebayo2018} in the input spatial domain, but, rather, it analyzes class activations in the tensor-state space.  

Finally, following the categorization of approaches to understanding community (group or cluster) structure in AI models presented by Watanabe et al. \cite{Watanabe2018}, our overarching approach to \emph{finding class encoding patterns} falls into the category 
``Analysis of trained layered neural networks''
and combines two subcategories: analysis of unit outputs and their mutual relationships and analysis of the influence on neural network inference by data.  Overall, our approach can be related to modular partitioning \cite{Hod2021}, \cite{Filan2021}, and unsupervised disentanglement of a learned representation \cite{Locatello2018}, \cite{Locatello2020}. 
In \cite{Hod2021}, \cite{Filan2021}, the authors search for local specializations of AI model subgraphs by using spectral clustering of AI model computation graphs and introducing two metrics, such as accuracy changes during neuron pruning ablation (neuron importance) and  class-specific accuracy drop in a subcluster of neurons (input-feature coherence). 
In contrast to \cite{Hod2021}, \cite{Filan2021}, our clustering of  computation units does not use "strong" and "weak" structural undirected connectivity of neurons as in spectral clustering, but, rather, repetitive co-occurrence patterns of utilization values in connected computation units. 
From the perspective of representation disentanglement, two studies by Locatello et al. \cite{Locatello2018}, \cite{Locatello2020} perform extensive experiments on $14\,000$ models trained on eight datasets to conclude that well-disentangled models cannot be identified without supervision and the evaluation metrics  do not always agree on what should be considered as “disentangled”. While we tacitly assume that high-dimensional data can be explained by lower dimensional semantically meaningful latent variables as by Locatello et al.  \cite{Locatello2018}, \cite{Locatello2020}, we do not attempt to fully automate finding subgraphs (i.e., a human is always in the loop) to follow the conclusions by Locatello et al. \cite{Locatello2018}, \cite{Locatello2020}. In addition, we do not aim at fully partitioning all AI model computation graphs into poly-semantic and mono-semantic subgraphs based on the poly-semantic and mono-semantic classification of neurons in subgraphs according to Olah et al. \cite{olah2020zoom} and Räuker et al. \cite{Rauker2022survey}.



\section{Methods}
\label{methods}

In this section, utilization-based class encodings are defined by addressing the three key challenges listed in Section~\ref{introduction}: (1) AI graph size and connectivity complexity, (2) component (graph node) heterogeneity, and (3) tensor dimensionality and real-value variability.

The utilization measurements of class encodings are defined by introducing tensor-states measured at the output of each component in AI computation graphs as training data points pass through the AI graph. The process of creating clean and poisoned training datasets is described next.

\begin{figure}
\includegraphics[
  width=12cm,
  height=6cm,
  keepaspectratio,
]{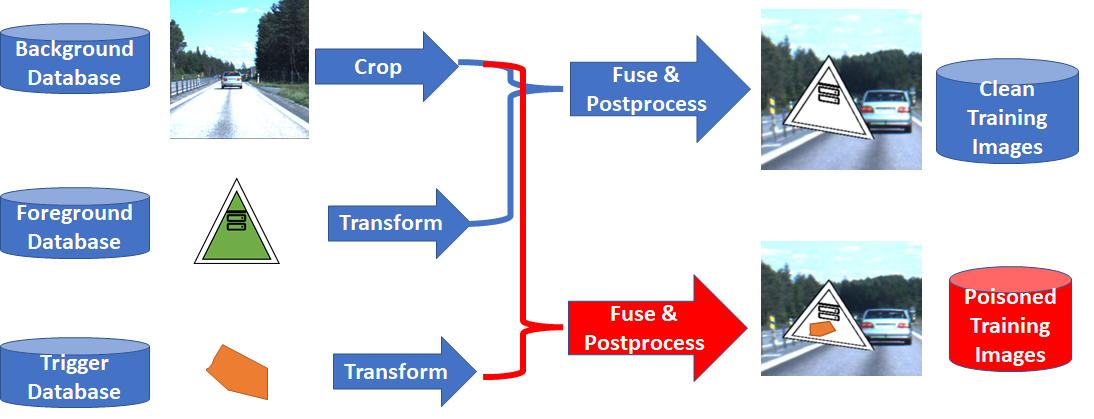}
  \centering
  \caption{The process of creating training data with traffic signs. An example of a simulated triangular traffic sign and a polygon type of trigger. }
  \label{fig:04x}
\end{figure}

\textbf{Creation of clean and poisoned training datasets:} 
The training images for each class in TrojAI challenge (Rounds 1-4) are created according to Figure~\ref{fig:04x} by fusing and post-processing foreground and background images. Images of foreground traffic signs are constructed from images of real and simulated traffic signs. The background images are retrieved from existing road and city video sequences (e.g., citiscapes \cite{Cityscapes2016},  KITTI 360 by Karlsruhe Institute of Technology and Toyota Technological Institute at Chicago \cite{Menze2015CVPR}, and others \cite{swedish_roads2022}). A variety of images per traffic sign class is accomplished by changing parameters of crop, transformation, fusion, and post-processing operations as shown in  Figure~\ref{fig:04x}.
  
\subsection{Utilization-based Class Encoding: Definitions}

\begin{figure}
\includegraphics[
  width=12cm,
  height=6cm,
  keepaspectratio,
]{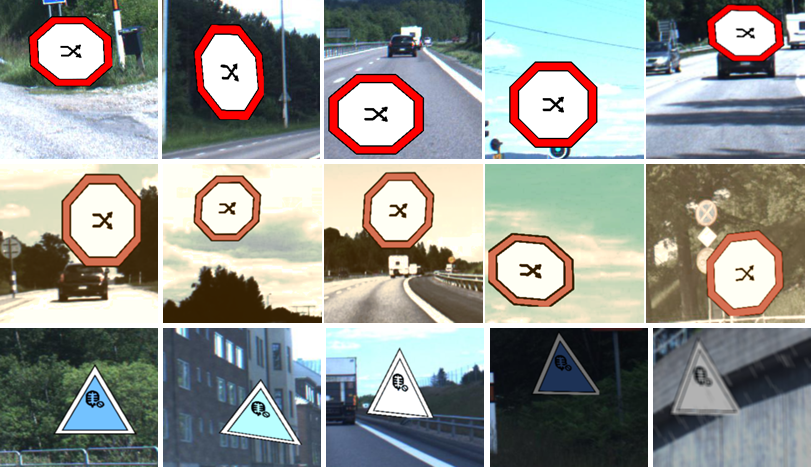}
  \centering
  \caption{Instagram trigger: Examples of clean source class images  $\vec{x}_{i}$ (top), poisoned source class images $g(\vec{x}_{i})$ with $g$ being a Kelvin Instagram filter (middle), and clean target class images $F_{a}(g(\vec{x}_{i}))=c_{t}$(bottom). }
  \label{fig:03a}
\end{figure}

\begin{figure}
\includegraphics[
  width=12cm,
  height=6cm,
  keepaspectratio,
]{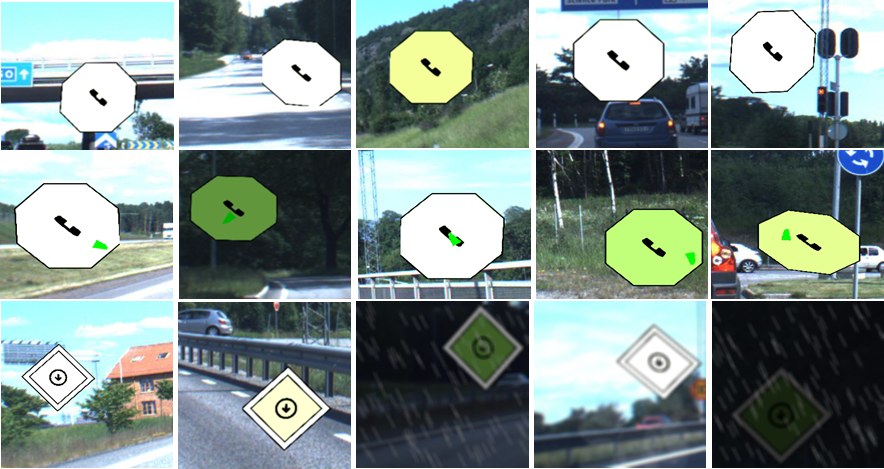}
  \centering
  \caption{Polygon trigger: Examples of clean source class images  $\vec{x}_{i}$ (top), poisoned source class images $g(\vec{x}_{i})$ with $g$ being a bright green polygon trigger (middle), and clean target class images $F_{a}(g(\vec{x}_{i}))=c_{t}$(bottom).}
  \label{fig:03c}
\end{figure}

\textbf{Clean and Poisoned AI models:} Let $F_{a}: \mathbb{R}^{m} \rightarrow \{1,..., C\}$ 
refer to a trained AI model with architecture $a$ that classifies two-dimensional $m$-variate images into one of $C$ classes.  When $F_{a}$ is clean (denoted as $F_{a}^{\square}$), $F_{a}$ achieves a high classification accuracy over input images 
$\vec{x}_{i} \in  \mathbb{R}^{m}; i \in \{1, ..., M\}$ where $M$ is the number of pixels. 
When a clean $F_{a}$ is poisoned by a trigger (denoted as $F_{a}^{\blacksquare}$), 
there exists a function 
$g:  \mathbb{R}^{m} \rightarrow \mathbb{R}^{m}$ applied to input images from a source trigger class $c_{s}$, such that $F_{a}(g(\vec{x}_{i}))=c_{t}$, where $c_{t}$ is the target trigger class and $c_{t} \neq c_{s}$. 
Examples of clean images from source class, poisoned images from source class, and clean images from target class are shown in Figure~\ref{fig:03a} (Instagram filter trigger) and in Figure~\ref{fig:03c} (Polygon trigger).  For a pair of trained clean and poisoned AI models, labels for source class $c_{s}$ and target class $c_{t}$ are predicted with high accuracies according to the four equations below:

\begin{equation}
F_{a}^{\square}(\vec{x})=c_{s} \;\; \textrm{and} \;\; F_{a}^{\square}(g(\vec{x}))=c_{s}
 \label{eq:00a}
\end{equation}
\begin{equation}
F_{a}^{\blacksquare}(\vec{x})=c_{s} \;\; \textrm{and} \;\;  F_{a}^{\blacksquare}(g(\vec{x}))=c_{t}
 \label{eq:00b}
\end{equation}


\textbf{AI computation graph:} A computation graph of a trained AI model $F_{a}$ is denoted by $G_{a} = \{V, E\}$ where $V = \{v_{1}, v_{2}, . . . , v_{n(a)}\}$ are the $n(a)$ computation units (or graph nodes or graph components) and $E \subseteq V \times V$ are the edges. The unidirectional edges of a graph $G_{a}$ are described by an adjacency matrix $A \in \{0, 1\}^{n(a) \times n(a)}$ with $A_{ij} = 1$ for all connected nodes $v_{i}$ and $v_{j}$, and $A_{ij} = 0$ for all other node pairs.

\begin{figure}
\includegraphics[
  width=12cm,
  height=6cm,
  keepaspectratio,
]{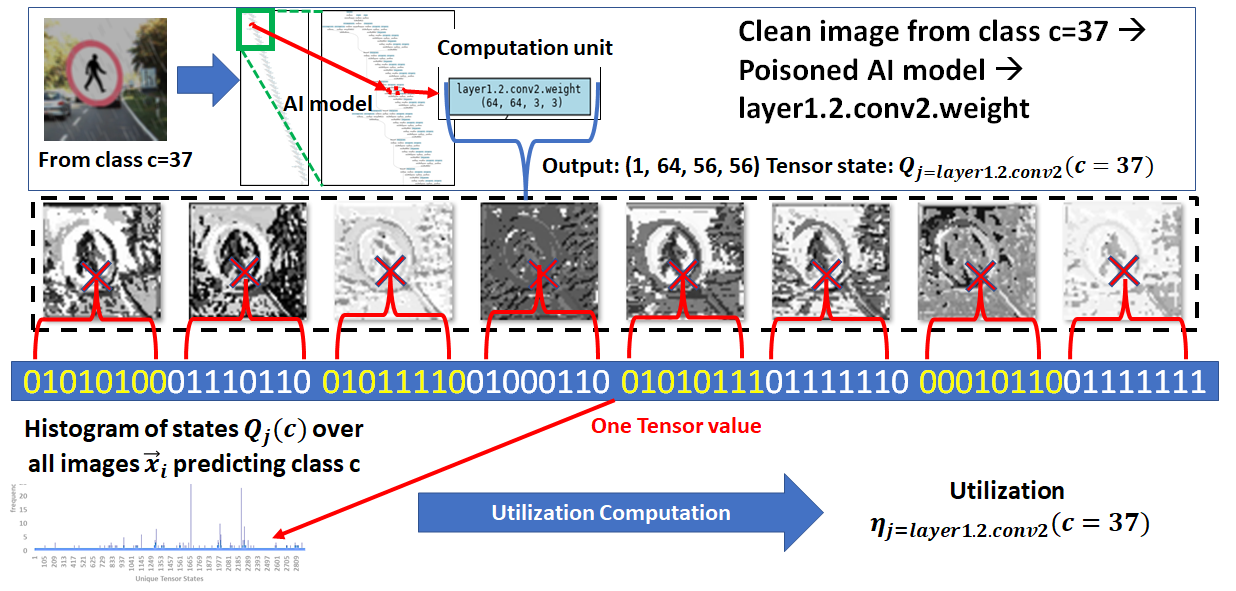}
  \centering
  \caption{An example of a tensor-state and its one value derived from a ResNet101 computation graph (node layer1.2.conv.weight) contributing to a tensor-state distribution used for utilization computations.}
  \label{fig:03d}
\end{figure}

\textbf{tensor-state:} Each input image $\vec{x}_{i}$ passes through $G_{a}$ populated with trained coefficients. The input generates a tensor of output values at each computation unit (i.e., an activation map) 
$v_{j}: \mathbb{R}^{D_{j}^{In}} \rightarrow \mathbb{R}^{D_{j}^{Out}}$, where $D_{j}^{In}$ and $D_{j}^{Out}$ are the input and output dimensions of data at the computation unit $j$. 
The output values are binarized by zero value thresholding to form a tensor-state 
$s_{j}(\vec{x}_{i})=b(v_{j}(\vec{x}_{i})) \in \{0,1\}^{D_{j}^{Out}}; b:\mathbb{R}^{D_{j}^{Out}} \rightarrow {\{0,1\}}^{D_{j}^{Out}}$. We refer to the graph location of $v_{j}$ at which the output values are measured as a probe location.  Figure~\ref{fig:03d} illustrates one tensor-state value for a specific ResNet101 computation graph, its specific graph node named layer1.2.conv2.weight, and one image from a predicted class $c=37$. The example tensor-state $(1,64,56,56)$ is visualized as a set of 8 images with dimensions $56 \times 56$ pixels, and the 64 bits (binarized outputs) are represented as 8 bytes. 

\textbf{tensor-state Distribution:} Given a set of measured tensor-states $\{s_{j}(\vec{x}_{i})\}$ at a computation unit $v_{j}$ for which $F_{a}(\vec{x}_{i}) = c$, 
let us denote $Q_{j}(c)=\{ q_{ij}(c) \}_{i=1}^{n_{j}}$ to be a discrete probability distribution function (PDF) over all tensor-state values, where $n_{j} = 2^{D_{j}^{Out}}$ is the maximum number of available tensor-state values at the $j$-th  computation unit $v_{j}$. The value of $q_{ij}(c)$ is the sum of counts of unique tensor-state values $count_{ij}$ invoked by all images $i$ ($\bigvee i \rightarrow s_{j}(\vec{x_{i}})$) and normalized by the maximum number of available tensor values $n_{j}$. 
Figure~\ref{fig:03d} (bottom left) shows the histogram values $count_{ij}$ computed from $5 \ 366 \ 576$ unique tensor-state values over all $2 \ 500$ training images of \emph{STOP pedestrian crossing} traffic signs. Based on the tensor-state dimensions $(1,64,56,56)$, one can establish the maximum number of predicted classes for such a node to be 
$C_{layer1.2.conv2}^{MAX}=\frac{2^{64}}{56*56*2500}  \approx 2.35*10^{12}$; a terascale count of traffic sign classes. 

\textbf{Reference tensor-state Distribution:} For a class-balanced training dataset with similar class complexities, let us refer to $P_{j} = \{ p_{ij} \}_{i=1}^{n_{j}}$ as the uniform (reference) PDF over all states; $p_{ij}=\frac{1}{2^{D_{j}^{Out}}}$.  The probabilities $p_{ij}$ are associated with each state (index $i$) and each  computation unit (index $j$) for each class $c$. 

\textbf{Utilization:} We can compute a scalar utilization value 
$\eta_{j}(c)$ for each class label $c$ and a computation unit $v_{j}$ from the count of measured states $q_{ij}(c)$ and the state distribution $Q_{j}(c)$ based on 
Equations~\ref{eq:01}-\ref{eq:03}. Equation~\ref{eq:01} defines utilization $\eta_{j}^{state}$ based on a deterministic view of states. In contrast, Equations~\ref{eq:01} and \ref{eq:02} define utilizations $\eta_{j}^{H}$ and $\eta_{j}^{KLDiv}$ based on a probabilistic view of states by computing entropy $H(Q_{j})$ of a state distribution normalized by maximum entropy $H_{j}^{max}$ or reference distribution $P_{j}$.    
 The three utilization definitions yield value ranges 
 $\eta_{j}^{state} \in [0,1]$,  $\eta_{j}^{H} \in [0,1]$, and $\eta_{j}^{KLDiv} \in [0,\infty]$ per  computation unit with an index $j$. For increasing utilization, the state- and entropy-based measurements will increase while the Kullback–Leibler(KL) Divergence-based measurement will decrease since it measures non-utilization (or a deviation from the reference uniform distribution of tensor-states across all predicted classes). The KL Divergence-based measurement assumes that the maximum number of available states $n_{j}$ is uniformly divided across all predicted classes (i.e., class encodings consume an equal number of available tensor-states).

\begin{equation}
	\eta_{j}^{state} = \sum_{i=1}^{n_{j}} \frac{count_{ij}}{ n_{j}} = \sum_{i=1}^{n_{j}} q_{ij} \leq 1
\label{eq:01}
\end{equation}

\begin{equation}
\eta_{j}^{H}=\frac{H(Q_{j})}{H_{j}^{max}}=
\frac{ -\sum_{i=1}^{n_{j}}( q_{ij}*\log_{2} { q_{ij} })}{ \log_{2} {n_{j}} } 
\label{eq:02}
\end{equation}

\begin{equation}
\eta_{j}^{KLDiv}=D_{KL}(Q_{j} \parallel P_{j})=\sum_{i=1}^{n_{j}}(q_{ij}*\log_{2} {\frac{q_{ij}}{p_{ij}}})
\label{eq:03}
\end{equation}

\begin{figure}
\includegraphics[
  width=12cm,
  height=6cm,
  keepaspectratio,
]{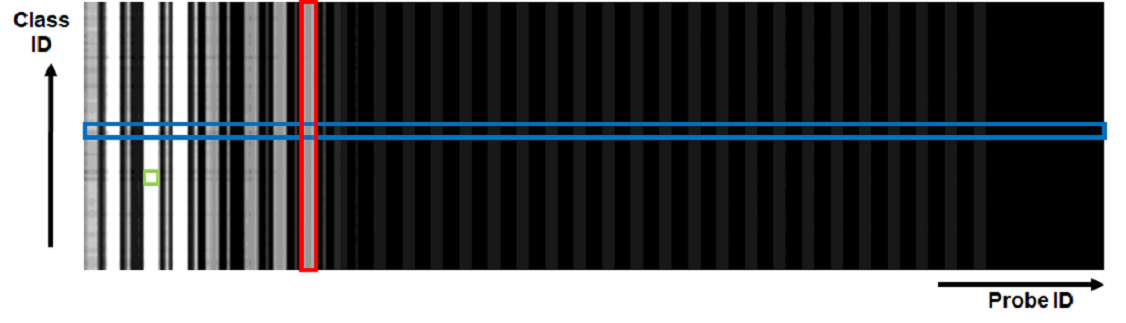}
  \centering
  \caption{AI model utilization fingerprint  $\mathbf{U_{a}}$ for a trained poisoned architecture $a$, ResNet101, predicting 75 classes ($C=75$) and the source class $c=25$ being poisoned by an Instagram filter (shown in Figure \ref{fig:03a}) to misclassify traffic signs to a target class $c=34$. The blue rectangle along a row shows a utilization-based class encoding $\vec{e}(c)$, the red rectangle along a column shows a utilization-based multi-class probe encoding $\vec{r}(j)$, and the small green square shows one utilization value $\eta_{j}(c)$. }
  \label{fig:03b}
\end{figure}

The vector of utilization values for all AI computation units $j \in \{1, ..., n(a)\}$ is referred to as \emph{a class encoding $\vec{e}(c)$ for the class $c$}. 
The vector of utilization values from all classes $c \in \{1,..., C\}$ if referred to as \emph{a probe encoding $\vec{r}(j)$ for the  computation unit $j$}.
A set of class encodings for $c \in \{1, ..., C\}$ ordered by the class label is denoted as \emph{an AI model utilization fingerprint $\mathbf{U_{a}}=\{\vec{e}(c=1), ..., \vec{e}(c=C)\}$}. An example of AI model fingerprint is shown in Figure~\ref{fig:03b} for ResNet101 architecture.

\textbf{Utilization Properties:} Utilization values are nondecreasing for increasing number of training data,  number of predicted classes, decreasing AI model capacity. Experimental verifications of these utilization measurement properties can be found in Appendix~\ref{properties}.

\subsection{Class Encoding: Measurements}

\begin{figure}
\includegraphics[
  width=12cm,
  height=6cm,
  keepaspectratio,
]{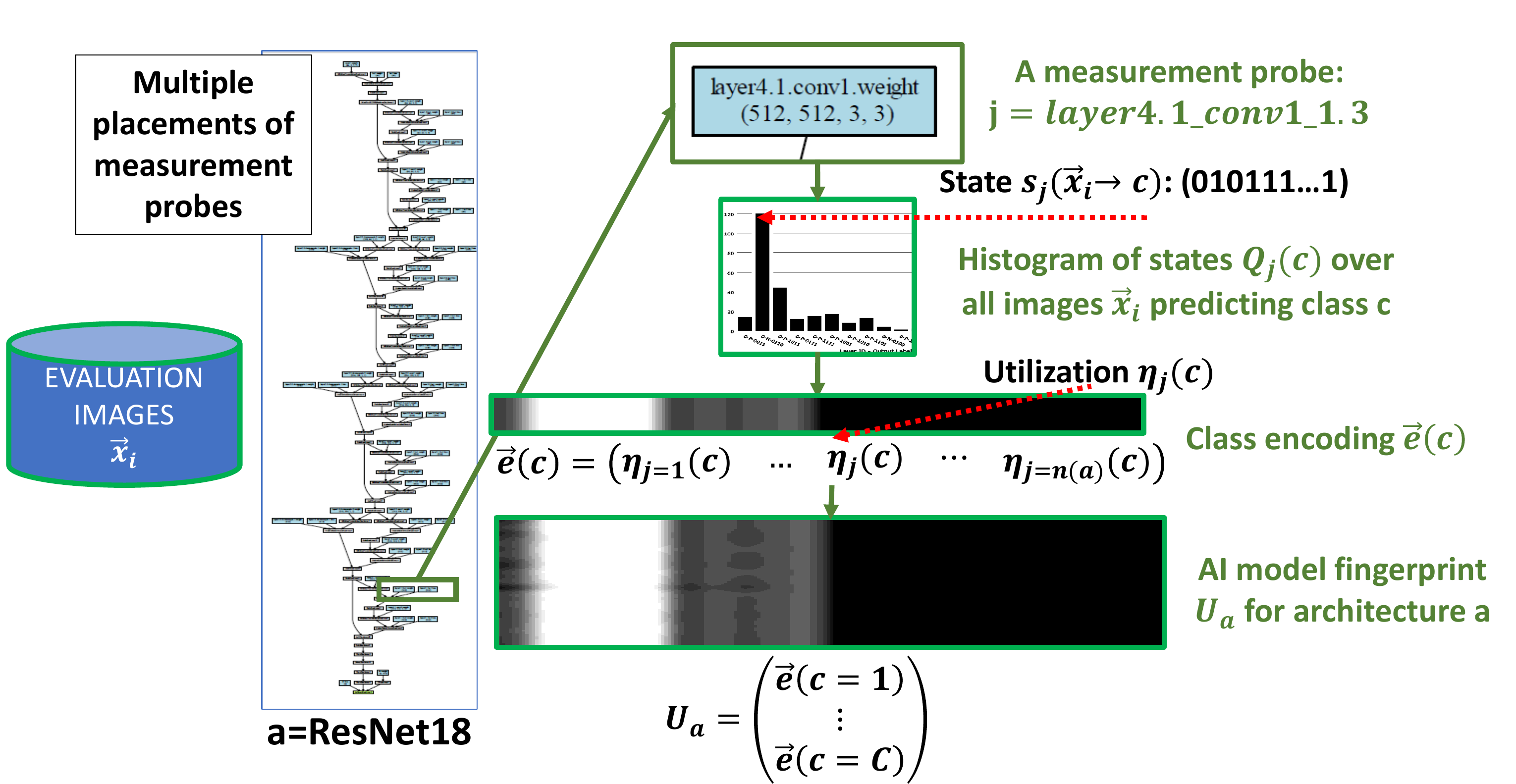}
  \centering
  \caption{An example workflow for computing an AI model fingerprint for ResNet18 architecture from a set of evaluation images.}
  \label{fig:03}
\end{figure}

\textbf{Utilization Measurement Workflow:}
Following the theoretical definition, the utilization workflow steps are shown in Figure \ref{fig:03}. The workflow starts with placing multiple measurement probes to collect the activation maps and follows the sequence of steps on the right side of Figure \ref{fig:03}: record tensor-states, compute a histogram of tensor-states, derive class encoding for one class, and form an AI model utilization fingerprint. The placement of a measurement probe is after each computation unit.   
 
\textbf{Computational Complexity of Utilization Measurements:}
\label{method:complexity}
 Following on the key challenges introduced in Section~\ref{introduction}, the utilization measurements require significant computational resources for managing tensor-states in memory. The measurement involves building state histograms, computing the utilization values according to 
Equations~\ref{eq:01} - \ref{eq:03}
per  computation unit of AI computation graph, and repeating the calculations over hundreds of  computation units per graph while evaluating hundreds of thousands of images per AI model and thousands of trained AI models. The requirements to compute one utilization-based AI model fingerprint $\mathbf{U_{a}}$ can be estimated based on Equations \ref{eq:04} (execution time $T(\mathbf{U_{a}}) $) and \ref{eq:05} (Memory).

\begin{equation}
T(\mathbf{U_{a}}) = M \times \widehat{T}(F_{a}(\vec{x}_{i}))
\label{eq:04}
\end{equation}
\begin{equation}
\textrm{Memory} \leq \max_{j}(D_{j}^{Out}) \times M \times n(a) 
\label{eq:05}
\end{equation}

where $M$ is the number of input images for all $C$ predicted classes,
$\widehat{T}(F_{a}(\vec{x}_{i}))$ is the estimated average inference time per image,
$\max_{j}(D_{j}^{Out})$ is the maximum output cardinality of a  computation unit $v_{j}$ (max number of output nodes),
and
$n(a)$ is the number of measurement probes in a graph for architecture $a$. For numerical examples, see Section~\ref{discussion}.

 We approached the computational challenges by 
 \begin{itemize}
\item reducing the number of training images per class and building an extrapolation model,
\item analyzing the AI model architecture designs to limit the number of probes, and 
\item modifying the KL Divergence computation according to \cite{bajcsy2021} to reduce computations. 
\end{itemize}
First, reducing the number of training images per class can be achieved via sampling and extrapolation modeling assuming that the classes are well represented in the tensor-state space with fewer samples (see Section~\ref{experimental_results}). 

\begin{figure}
\includegraphics[
  width=12cm,
  height=6cm,
  keepaspectratio,
]{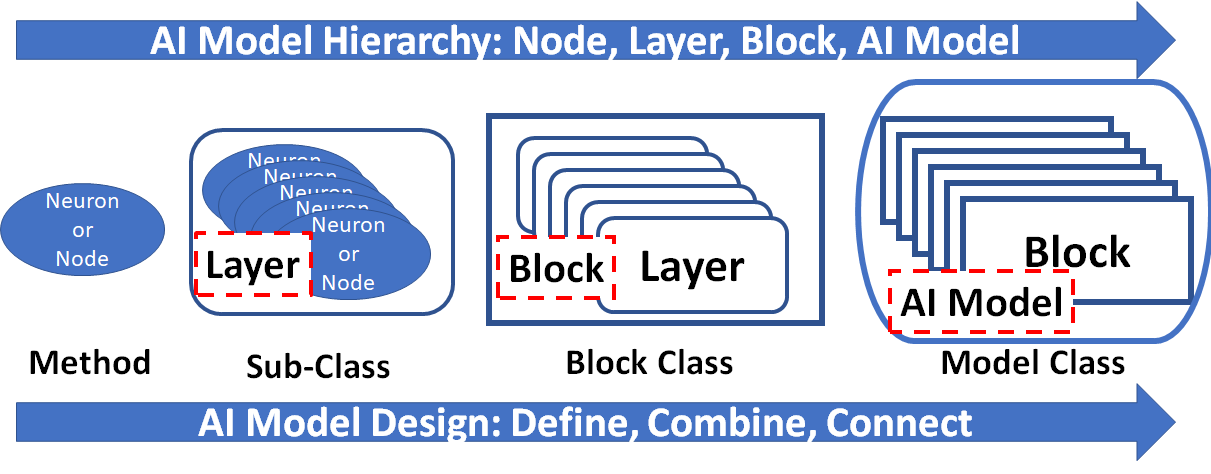}
  \centering
  \caption{A hierarchical design of complex AI models as a logical partition of AI model computation graphs.}
  \label{fig:04a}
\end{figure}

Second, one can leverage a hierarchical structure of some AI computation graphs. Figure~\ref{fig:04a} shows a hierarchy of nodes, layers, and blocks that form an AI model. 
As AI model architecture designers define, combine, and connect computation units, AI computation graphs are partitioned into programming modules (i.e., methods, layers or blocks) that can be used for placing measurement probes.  Since the programming modules represent a logical partition of a computation graph based on ad-hoc or exhaustive experimentations (i.e., the Neural Architecture Search problem \cite{nasbench2019}), they can define initial placements of measurement probes and lower memory and computational requirements for utilization computations. 

Third, the KL Divergence computation in Equation \ref{eq:03} requires 
aligning measured and reference states, which is computationally expensive (for corner cases: if $q_{ij}=0$ then $\eta_{j}^{KLDiv} = 0$ and if $p_{ij}=0$, then $\eta_{j}^{KLDiv} = 0$ because $\lim_{x \to 0} (x * \log_{2}x) = 0$).   We assume that, on average, the uniform (reference) PDF is equally divided among predicted classes (efficient class encodings), which eliminates  the need for aligning.


\textbf{Class-Specific Utilization Evaluation:}
Utilization values can be measured by evaluating clean and poisoned trained AI models by clean and poisoned training images. We considered utilization measurements to be derived from the sets described in Table~\ref{table:03}. Other evaluation options will be explored in the future.

\begin{table}
  \caption{Sets of utilization measurements}
  \label{table:03}
\centering
\begin{tabular}{|c | c | c | }
\hline
Data Set Index & AI Model & Evaluated with Class Images \\
\hline
Set 1 & Clean & Clean \\
\hline
Set 2 & Poisoned & Clean \\
\hline
Set 3 & Poisoned & Source class + trigger \\
\hline
\end{tabular}
\end{table}

\subsection{Utilization-based Class Encoding: Finding Patterns}

The problem of finding patterns is defined with respect to three granularity levels of AI models summarized in Table~\ref{table:03b}. The granularity levels are introduced to cope with the complexity of class encodings in AI models at the level of (a) individual computation units that encode tensor-states of images $s_{j}(\vec{x}_{i} \rightarrow c)$, (b) computation subgraphs that capture utilization motifs per class $\vec{e}(c)$, and (c) AI model fingerprints $U_{a}$ that represent utilizations of all classes in a computation graph. The micro to macro granularity levels can be leveraged for hierarchical analyses of AI models as we are inspecting thousands of AI models in TrojAI challenge rounds, predicting 15 to 45 classes, using 5 to 20 computation graphs (i.e., architectures), and $2500$ training images per class (see Table~\ref{table:01} listing complexity challenges).

\begin{table}
  \caption{Granularity of finding patterns}
  \label{table:03b}
\centering
\begin{tabular}{|c | c | c | }
\hline
\textbf{Granularity} & \textbf{Measurements} & \textbf{Analyzed metrics} \\
\hline
\makecell{Computation unit\\ (graph node)} & \makecell{Unique tensor-states\\ per class} & \makecell{Distribution of\\ common tensor-states} \\
\hline
\makecell{computation subgraph} & \makecell{Utilization class-\\ encoding vector} & \makecell{Vector correlations\\ of class encodings} \\
\hline
\makecell{computation graph\\ (fingerprint of AI model)} & \makecell{Utilization matrix\\ (class vs. probe)} & \makecell{Delta of utilization\\ histogram bins} \\
\hline
\end{tabular}
\end{table}

Table~\ref{table:03b} also lists the measurements and metrics at each granularity level. Measurements consist of unique tensor-states, utilization-based class encoding vectors, and utilization matrices. Metrics are applied to individual measurements or pairs of measurements to identify patterns, for instance, 
\begin{itemize}
\item spatial overlaps of semantically meaningful image regions with tensor-state values (e.g., common blue sky versus class-unique traffic sign symbols in invoked tensor-states), 
\item partial similarities of multiple class encodings in the same AI model (e.g., encoding of traffic sign classes utilizing similar and dissimilar AI model computation subgraphs), and 
\item similarity of AI model utilization fingerprints in AI model collections (e.g., common utilization of multi-class encodings in multiple AI model architectures).
 \end{itemize}
 
\textbf{Patterns detected in computation units:}
In addition to the computational challenges associated with computing utilization as described in Section~\ref{method:complexity}, one must address the visualization challenges for viewing multidimensional tensors. The challenge was approached by forming  8 images for one tensor-state of size $(1, 64, 56, 56)$ in Figure~\ref{fig:03d} where the tensor-state was invoked by one input image at one of the ResNet101 computation units. Equation~\ref{eq:05a} provides the formula for calculating a total number of images representing tensor-states that could be visually inspected.

\begin{equation}
N_{\textrm{Images}}^{\textrm{Tensors}}  = \sum_{j}^{n(a)}(\frac{D_{j}^{Out}}{8}) \times M 
\label{eq:05a}
\end{equation}

where $M$ is the number of input images for all $C$ predicted classes,
$D_{j}^{Out}$ is the number of outputs from a  computation unit $v_{j}$, and
$n(a)$ is the number of measurement probes in a graph for architecture $a$. 
For example, for the ResNet101 computation graph with $n(a)=286$ utilization measurement probes and $M = 100 \ 000$ images (40 predicted traffic sign classes represented by $2500$ training images per class) and on average $D_{j}^{Out}=64$ dimensional outputs, one would need to inspect about $286 \times \frac{64}{8} \times 100 \ 000 = 2.288*10^{8}$ grayscale images. 

\begin{figure}
\includegraphics[
  width=12cm,
  height=6cm,
  keepaspectratio,
]{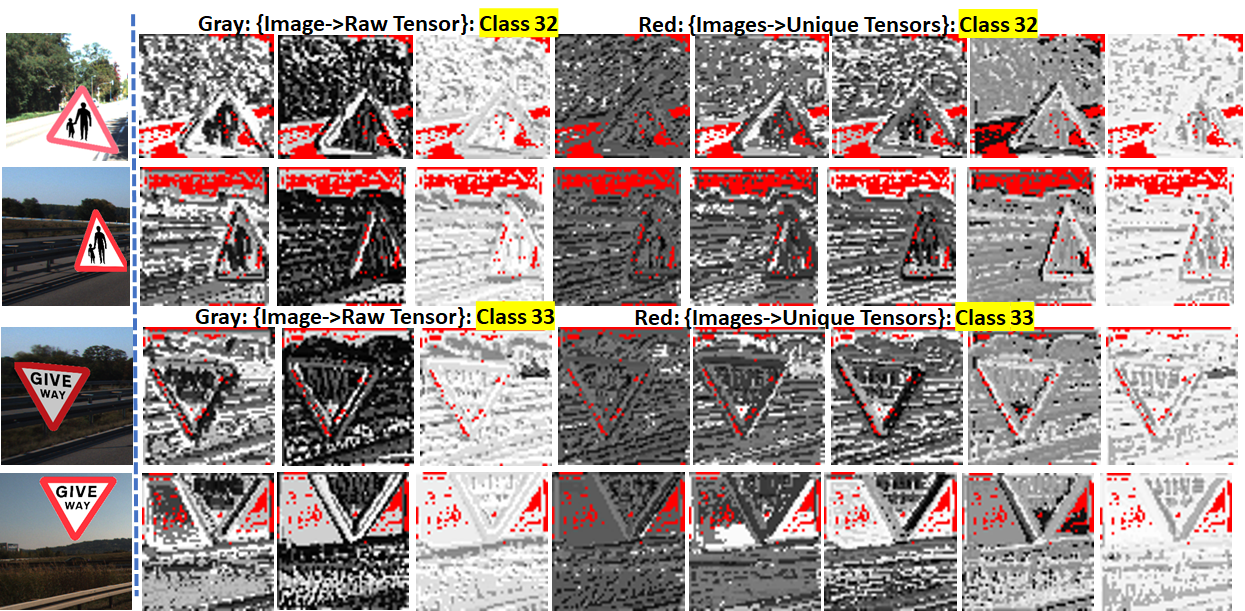}
  \centering
  \caption{Visualization of tensor-state values in red for two images from two classes (Class 32 - \emph{Parent and child} and Class 33 - \emph{GIVE WAY} road signs) in layer1.2.conv2 of the ResNet101 model that occur more than 100 times in 2500 training images of the same class. 
  }
  \label{fig:ts01}
\end{figure}

To lower the number of images, we focus primarily on pairs of classes (clean and clean or clean and its corresponding poisoned classes). To simplify our visual inspection, we look for all unique tensor-state values in all training images per traffic sign class with a frequency higher than a threshold. These high-frequency tensor-state values are then highlighted in each set of 8 images representing one tensor-state invoked by one training image as shown in Figure~\ref{fig:ts01}. Based on images like in Figure~\ref{fig:ts01}, one can derive conclusions about the presence of features in each image that are common across a training collection defining a class. For instance, Figure~\ref{fig:ts01} (top rows) would indicate that the class  \emph{Parent and child} road signs contains many blue sky and saturated regions, as well as some key discriminating parts within the triangular road sign.

\begin{figure}
\includegraphics[
  width=12cm,
  height=6cm,
  keepaspectratio,
]{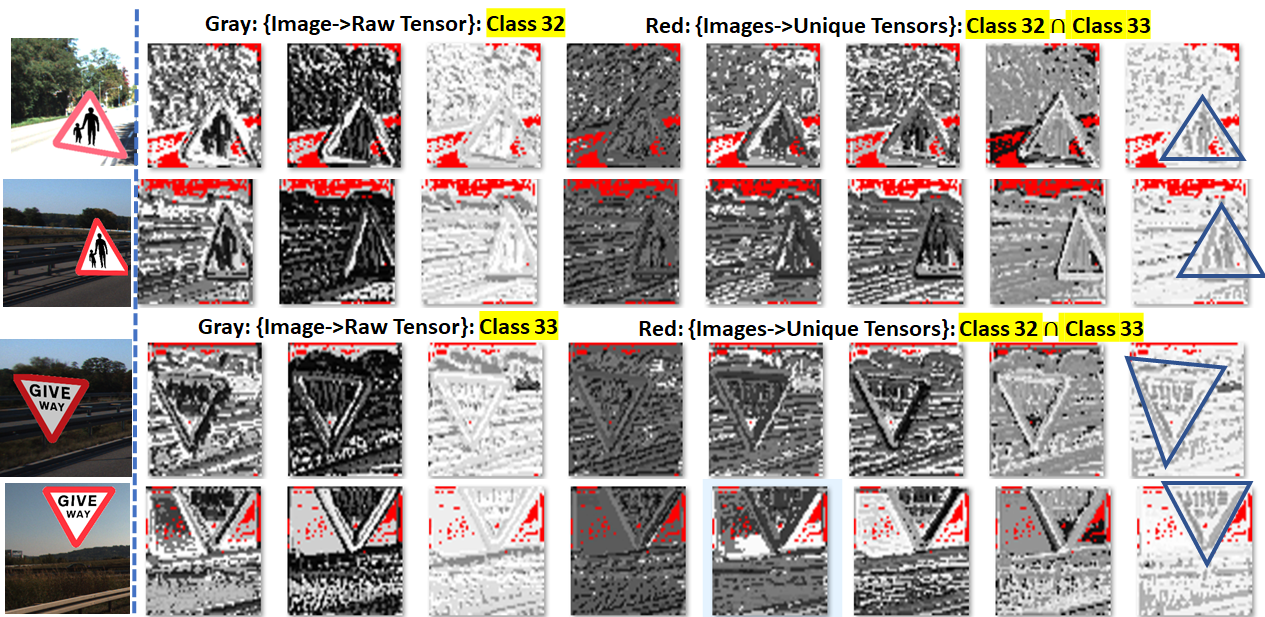}
  \centering
  \caption{Visualization of tensor-state values in red for two images from two classes (Class 32 - \emph{Parent and child} and Class 33 - \emph{GIVE WAY} road signs) in layer1.2.conv2 of ResNet101 that occur more than 100 times in 2500 training images of one class and are present in two example images of the other class. 
  }
  \label{fig:ts01b}
\end{figure}

Figure~\ref{fig:ts01b} shows two images from class 32 with red dots overlaid at tensor-state values overlapping with class 33 (top two rows). The bottom two rows illustrate two images from class 33 with red dot overlaid at tensor-state values characterizing all images in class 32. By comparing Figure~\ref{fig:ts01} and Figure~\ref{fig:ts01b}, one can observe that both classes (1) have common tensor-state values corresponding to the blue sky and saturated regions and (2) do not overlap in tensor-state values defining the foreground traffic signs (\emph{Parent and child} and \emph{GIVE WAY}) except from a small number of white pixels. Note that a few tensor-state values in the red rim of each traffic sign are common to each class according to Figure~\ref{fig:ts01}, but they are not common to both classes as they differ in the shades of red.   
Following the classification of graph computation units as poly-semantic and mono-semantic neurons \cite{olah2020zoom}, \cite{Rauker2022survey}, we can also conclude that \emph{layer1.2.conv2} is poly-semantic since it is constructing common  and unique class characteristics in the two traffic sign classes and passing them to the downstream graph computation units.

\begin{figure}
\includegraphics[
  width=12cm,
  height=6cm,
  keepaspectratio,
]{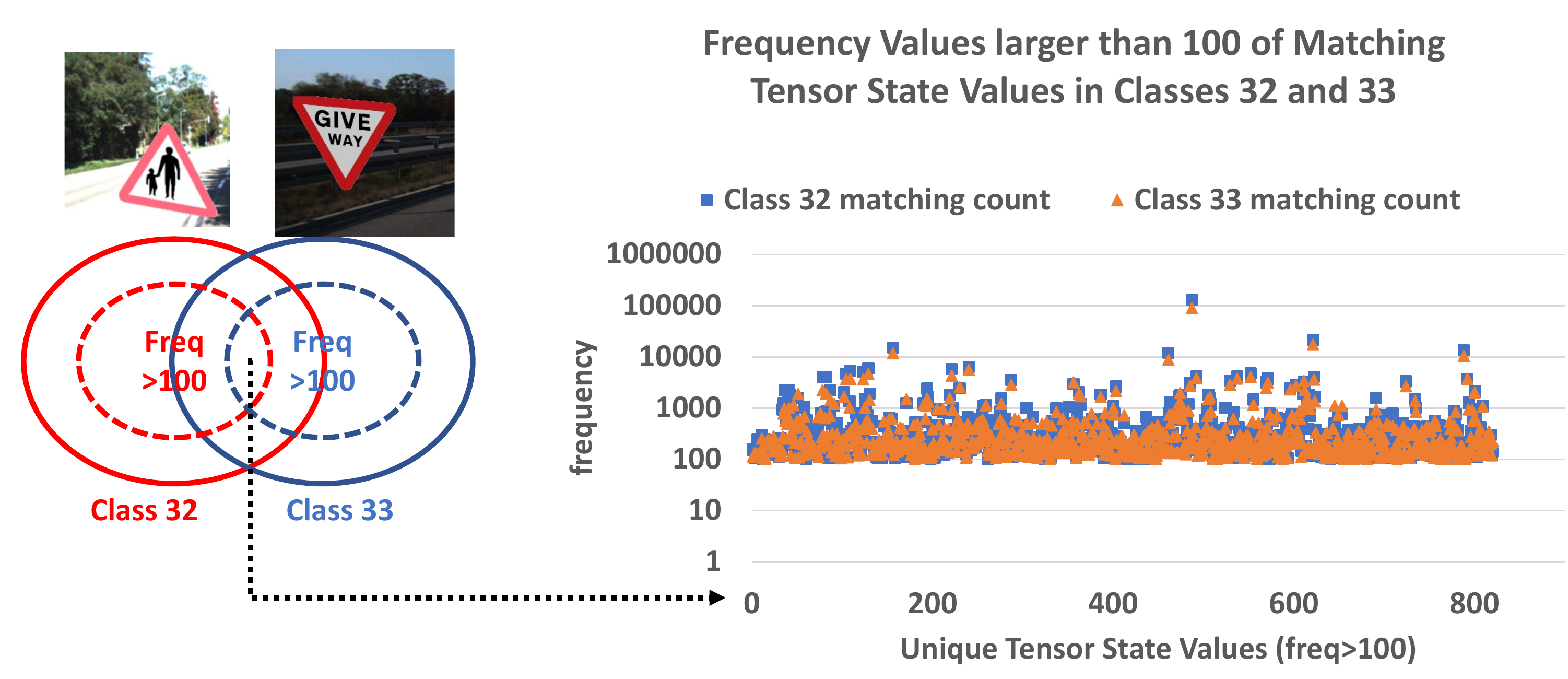}
  \centering
  \caption{Left - Venn diagram of images and their unique tensor-states from two classes. The dashed circles refer to subsets of unique tensor-states that are invoked at least 100 times by all training images from each class. Right - Histograms of unique tensor-state values in layer1.2.conv2 of ResNet101 that occur more than 100 times in 2500 clean images shown in Figure~\ref{fig:ts01}. The dotted line connects the unique tensor-states in the histogram with the Venn diagram intersection region. 
  }
  \label{fig:ts02}
\end{figure}

Another approach to inspecting the class encodings is via histograms of tensor-state values. Table~\ref{table:04} summarizes statistics of the number of unique tensor-state values in layer1.2.conv2 of ResNet101 AI model for two classes of training images shown in Figure~\ref{fig:ts01}. To scale down the visualization requirements on a histogram with more than $5.8$ million bins, we can threshold the bins based on tensor-state value frequencies. Figure~\ref{fig:ts02} shows the histogram visualization for the threshold value equal to $100$ using Microsoft Excel. The frequency (count) along a vertical axis is shown on a logarithmic scale to accommodate the wide range of count values. The tabular and histogram visualizations allow to observe the number of tensor-state values, their frequencies, and overlapping characteristics of classes as illustrated in Figure~\ref{fig:ts02}. The histogram in Figure~\ref{fig:ts02}(right) illustrates how overlapping unique tensor-states between class 32 and class 33 with frequencies larger than 100 (horizontal axis) would have difference frequencies (vertical axis) in each of the classes defined by 2500 images per class. These overlaps can be explained by using the same background pool of images with characteristics on its own (e.g., blue sky, trees, roads).

\begin{table}
  \caption{Number of unique tensor-state values invoked by 2500 clean images and 2500 poisoned images with frequencies higher than 0, 1, 10, and 100 in layer1.2.conv2 of ResNet101 (image examples are shown in Figure~\ref{fig:ts01}).
)}
  \label{table:04}
\centering
\begin{tabular}{|c | c | c |}
\hline
Num. unique state values 	& Class 32 &	Class 33   \\
\hline
\makecell{$>0$} & \makecell{$5 \, 610 \, 111$} & \makecell{$5 \, 815 \, 399$}  \\
\hline
\makecell{$>1$} & \makecell{$508 \, 419$} & \makecell{$529 \, 384$} \\
\hline
\makecell{$>10$} & \makecell{$28 \, 606$} & \makecell{$25 \, 870$} \\
\hline
\makecell{$>100$} &  \makecell{$1476$} & \makecell{$3633$}  \\
\hline
\makecell{$\eta_{j=layer1.2.conv2}^{entropy}$} & \makecell{$33.2$} & \makecell{$33.6$}  \\
\hline
\end{tabular}
\end{table}

\begin{figure}
\includegraphics[
  width=12cm,
  height=6cm,
  keepaspectratio,
]{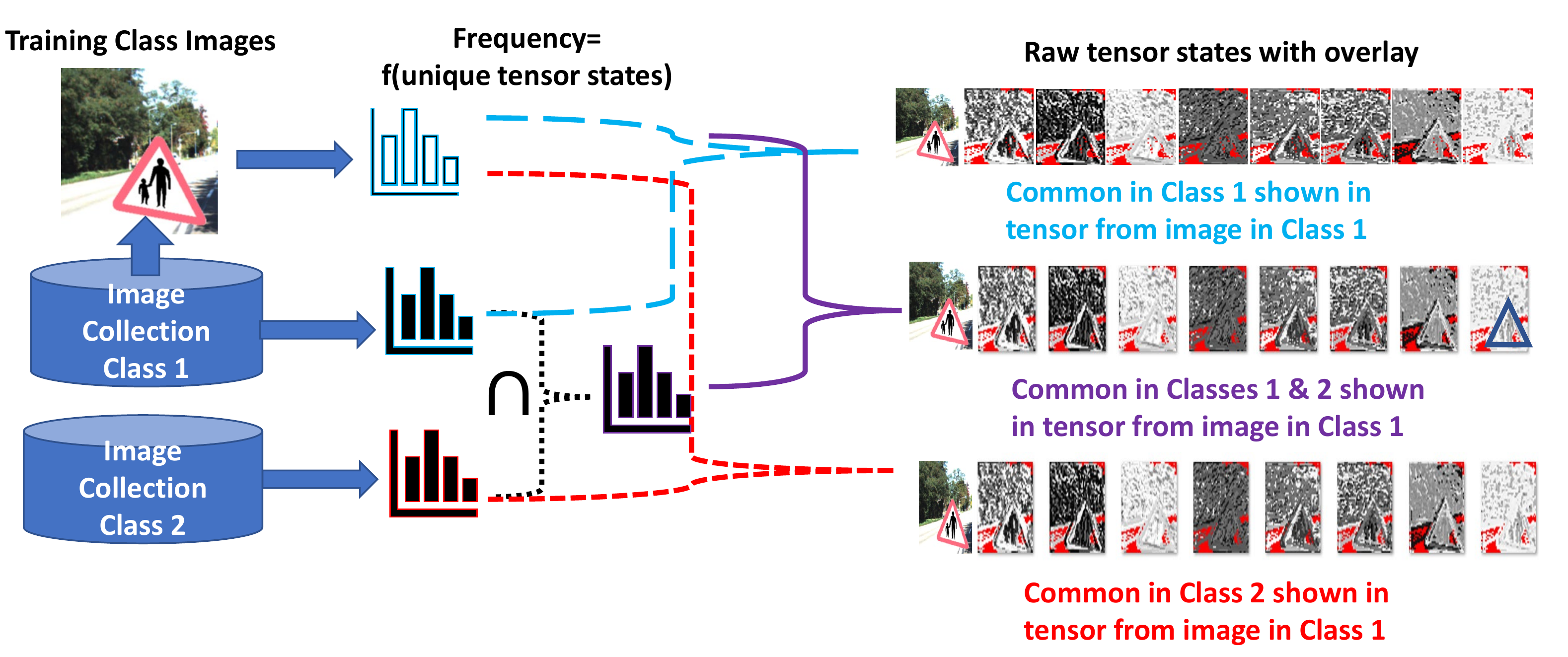}
  \centering
  \caption{Conceptual comparisons of presence or absence of tensor-state values across two traffic sign classes. The right column shows in red the intersections of unique tensor-states from a single image in class 1 with the tensor-states coming from (a) a collection of images in the same class 1 (top right), (b) the intersection of tensor-states measured from two image classes (middle right), and (c) a collection of images in the other class 2 (bottom right).  
  }
  \label{fig:ts03}
\end{figure}

Note that in order to understand unique and overlapping class characteristics one would need to compute a much more complex Venn diagram for an AI model predicting $C$ classes than the one in Figure~\ref{fig:ts02} (left). While common and distinct characteristics of two classes can be compared in three different ways as illustrated in Figure~\ref{fig:ts03}, the number of comparisons for $C$ classes would be $2^{C} - 1$, which quickly exceeds the limits of human inspections (e.g., for $C=40$ predicted traffic signs one would need to perform approximately $10^{12}$ comparisons).

\textbf{Patterns detected in computation subgraphs:}
The next level of granularity in explainable AI is to detect patterns in computation graphs according to the utilization-based class encodings as shown in Figure~\ref{fig:01} (right).  The motivation comes from the initial reports about semantically meaningful outputs of a group of computation units in AI models (e.g., partial curve detectors \cite{olah2020zoom}) and a hypothesis that such groups would repeat to encode more complex curves.  
The problem lies in finding subgraphs in an AI computation graph (AI model) that are utilized the same way. This problem is known to be NP-hard (non-deterministic polynomial-time hardness) \cite{Asahiro2014}.

\begin{figure}
\includegraphics[
  width=12cm,
  height=6cm,
  keepaspectratio,
]{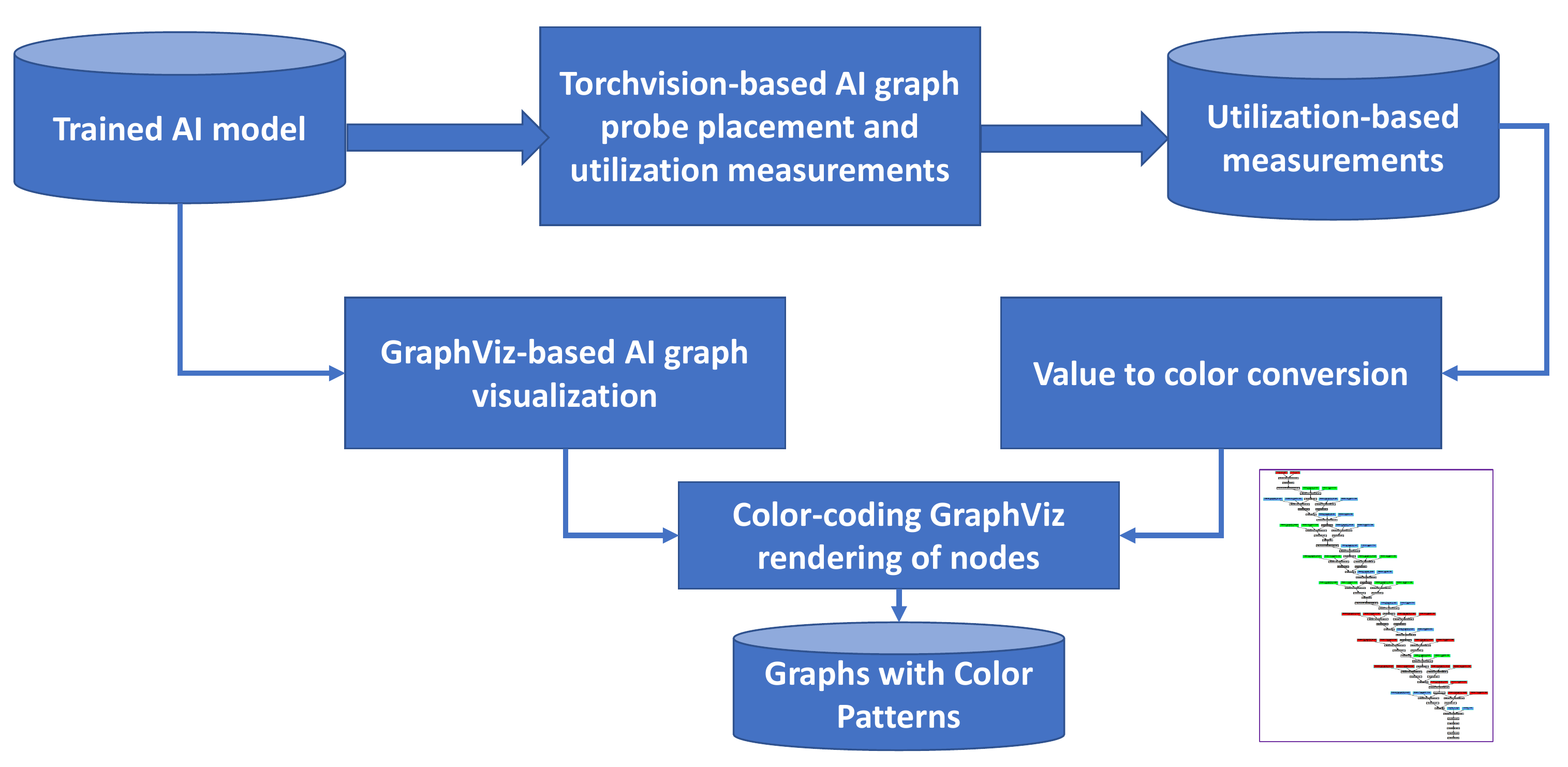}
  \centering
  \caption{A workflow for pseudo-coloring graph nodes according to utilization measurements.}
  \label{fig:04b}
\end{figure}

The baseline approach is to use a human visual inspection to identify class encoding patterns. Figure~\ref{fig:04b} illustrates the use of Torchvision \cite{torchvision2010}, \cite{torchvision2019} for placing the measurement probes and DiGraph \cite{digraph2019} for visualizing AI model computation graph with nodes color-coded according to the utilization values. 
Due to the visualization tradeoffs between rendered global and local information of very large graphs, complex connectivity, and heterogeneity of graph nodes, this visual inspection approach to finding patterns is not suitable for global comparisons of AI models (graphs).
  
\begin{figure}
\includegraphics[
  width=12cm,
  height=10cm,
  keepaspectratio,
]{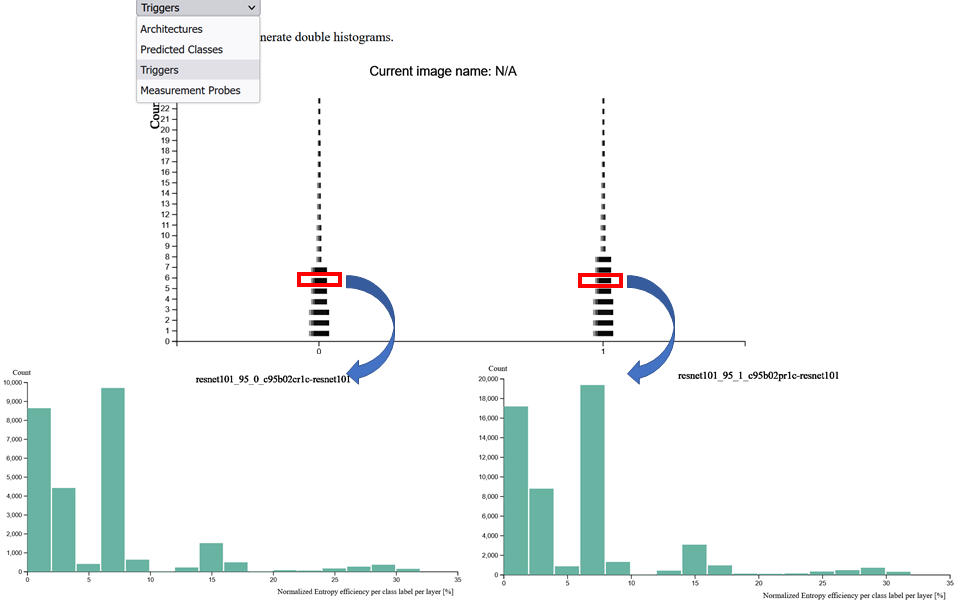}
  \centering
  \caption{A pair-wise web-based comparison of AI model fingerprints. Top - histogram of AI model fingerprints with the drop-down menu for sorting based on attributes (top left). Bottom - two histograms from selected AI model fingerprints shown in red boxes.}
  \label{fig:04z}
\end{figure}

\textbf{Patterns detected in computation graphs:}
A fingerprint visualization shown in Figure~\ref{fig:03b} is a coarse-level summary of class encodings per AI model architecture. While analyzing a single AI model, one can immediately see similarity and dissimilarity of class encodings among classes along the vertical axis and utilization patterns of  computation units in an AI computation graph along the horizontal axis. When analyzing multiple AI models, one can compare histograms of utilization values derived from multiple fingerprints. To support such comparisons, we have developed a web-based fingerprint comparison for collections of AI models that can scale to comparisons of thousands of AI models with different architectures (e.g., 1008 AI models with 16 architectures in Round 4 of TrojAI challenge). Model fingerprints can be sorted based on number of triggers, number of predicted classes, number of probes, and architecture types, and then compared via utilization histograms as illustrated in Figure~\ref{fig:04z}. 

\section{Experimental Results}
\label{experimental_results}

\subsection{Experimental Datasets}
 
We reused the software for generating the TrojAI Challenge datasets in Round 4 \cite{trojai_software2022}. 
Due to a large variability of training images used for training AI models in TrojAI Rounds 1 to 4 (the training datasets are different for each of the thousands of trained AI models per Round), we generated only two training datasets
with a fixed generation seed while varying other parameters. In addition, we increased the number of predicted classes to $C=75$ and $C=95$ traffic signs. Following Figure~\ref{fig:04x}, we created four training datasets (two training datasets $\times \{C=75, C=95\}$) and trained with them three architectures including VGG13, SqueezeNet v1.1, and ResNet101 (once without triggers (clean), once with Instagram filter triggers, and once with polygon triggers). 
One trigger was inserted into a poisoned training dataset with a constant trigger fraction between clean and poisoned images equal to $0.5$. Each AI model was trained three times with a different random training seed to explore the variability of class encodings.  Each class was represented by $2500$ color images, which amounted to $75 \times 2500 = 187\,500$ and $95 \times 2500 = 237\,500$ training images.
The entire dataset consisted of 108 trained AI models (training datasets $\times$ number of replicates $\times$ number of architectures $\times$ number of triggers $= 4 \times 3 \times 3 \times 3 = 108$).


\subsection{Finding Patterns by Comparing Clean and Poisoned Classes}

We describe the process of identifying graph locations for placing utilization measurement probes in Appendix~\ref{probe_placement}.
As mentioned before, utilization measurements satisfy multiple properties that are experimentally supported in Appendix~\ref{properties}. Similarly, we documented variability of utilization measurements in Appendix~\ref{variability} and computational requirements in Appendix~\ref{computational_requirements}.

To address the encoding complexity of clean and poisoned classes, we proceeded with the analyses from macro to micro granularity levels as documented in Table~\ref{table:03b}. We started with pattern detections in computation graphs first, next in subgraphs, and then in graph nodes. Our experiments are motivated by (a) evaluating our hierarchical utilization-based approach to classifying a large number of AI models and (b) understanding and validating the use of utilization measurements for this classification task at the tensor-state (micro) levels.

\subsubsection{Patterns detected in computation graphs:}
\label{pattern:graph}

We illustrate the utilization patterns in class encodings for four trained models in Round 4 holdout dataset of TrojAI challenge with the parameters summarized in Table~\ref{table:0421}. The type of parametrization, the number of parameters, and their wide ranges of values represent a large space of possible  trigger configurations. Such parametrizations are important for reducing a large search space of triggers by doing an experimental design for training poisoned AI models. 


\begin{table}
  \caption{Four trained models with the following parameters: architecture $a=$ ResNet101, 
number of predicted classes $C=17$, 
number of trojans $g_{i}(\vec{x})$ per AI model $\{0,1,2,2\}$, and trigger functions defined below.}
  \label{table:0421}
\centering
\begin{tabular}{|c | c | c | c |}
\hline
Model ID 	& Model Type & Trigger 0 & Trigger 1   \\
\hline
\makecell{$142$} & \makecell{Clean} & \makecell{$g_{0}(\vec{x})=\vec{x}$} & \makecell{$g_{1}(\vec{x})=\vec{x}$}  \\
\hline
\makecell{$235$} & \makecell{Poisoned} & \makecell{$g_{0}(\vec{x})=$ Kelvin filter} & \makecell{$g_{1}(\vec{x})=\vec{x}$}\\
\hline
\makecell{$150$} & \makecell{Poisoned} & \makecell{$g_{0}(\vec{x})=$ Gotham filter} & \makecell{$g_{1}(\vec{x})=$ Lomo filter} \\
\hline
\makecell{$250$} &  \makecell{Poisoned} & \makecell{$g_{0}(\vec{x})=$ 9-sided polygon \\of color $[200,0,0]$}  & \makecell{$g_{1}(\vec{x})=$ 4-sided polygon \\of color $[0,200,200]$}\\
\hline
\end{tabular}
\end{table}

In our study, all four AI models are evaluated with clean images (i.e., Set 1 in Table~\ref{table:03}).
The four AI models are trained with different traffic signs, assigned randomly to 17 classes, and placed on top of randomly chosen backgrounds from cityscapes, kitti road, and kitti city image collections, and, therefore, the fingerprints cannot be compared by element-to-element.

\begin{figure}
\includegraphics[
  width=12cm,
  height=6cm,
  keepaspectratio,
]{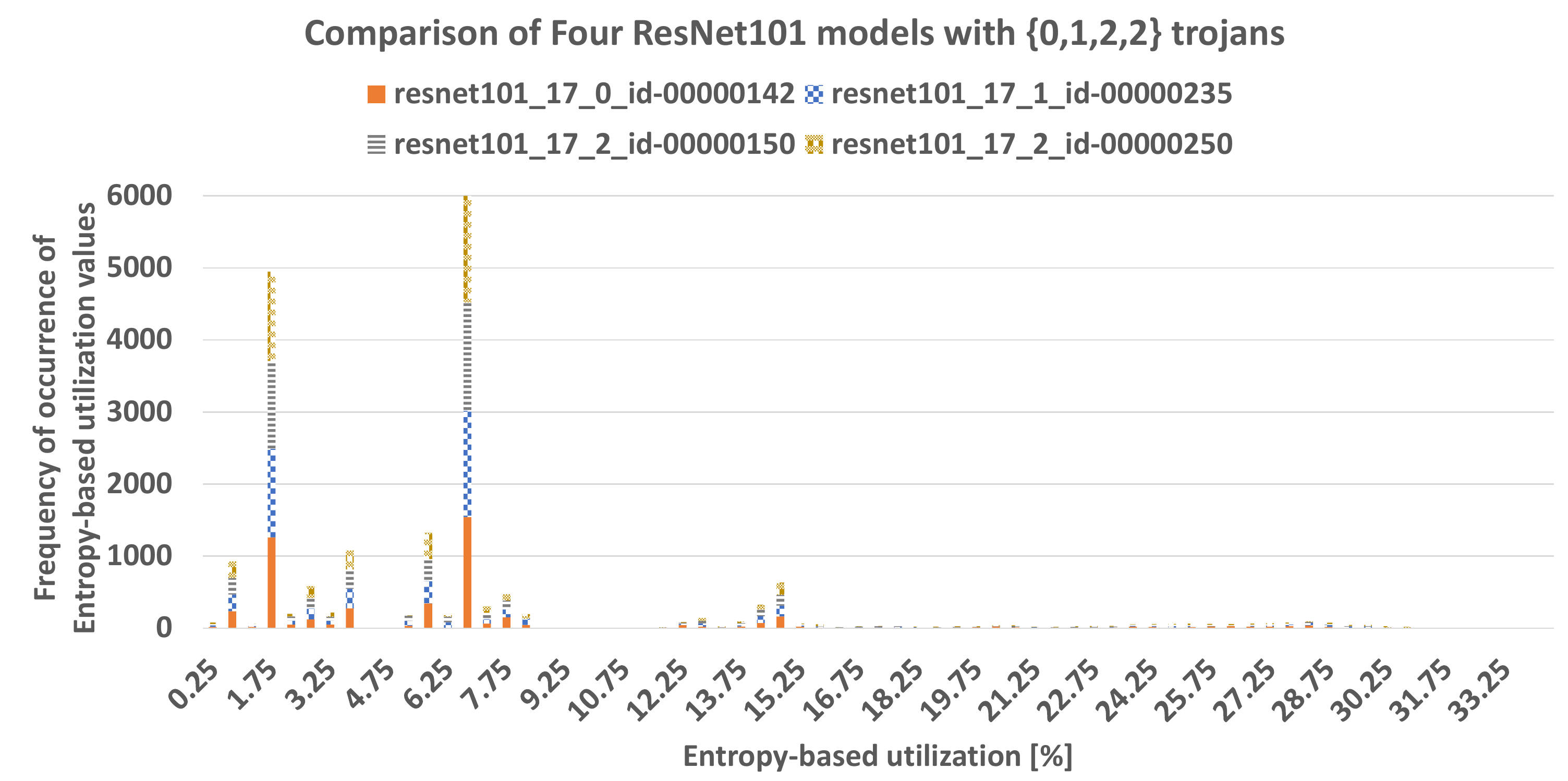}
  \centering
  \caption{Stacked histogram of four ResNet101 models with zero, one, and two triggers (one model with two Instagram triggers and one with model two polygon triggers). }
  \label{fig:15a}
\end{figure}
\begin{figure}
\includegraphics[
  width=12cm,
  height=6cm,
  keepaspectratio,
]{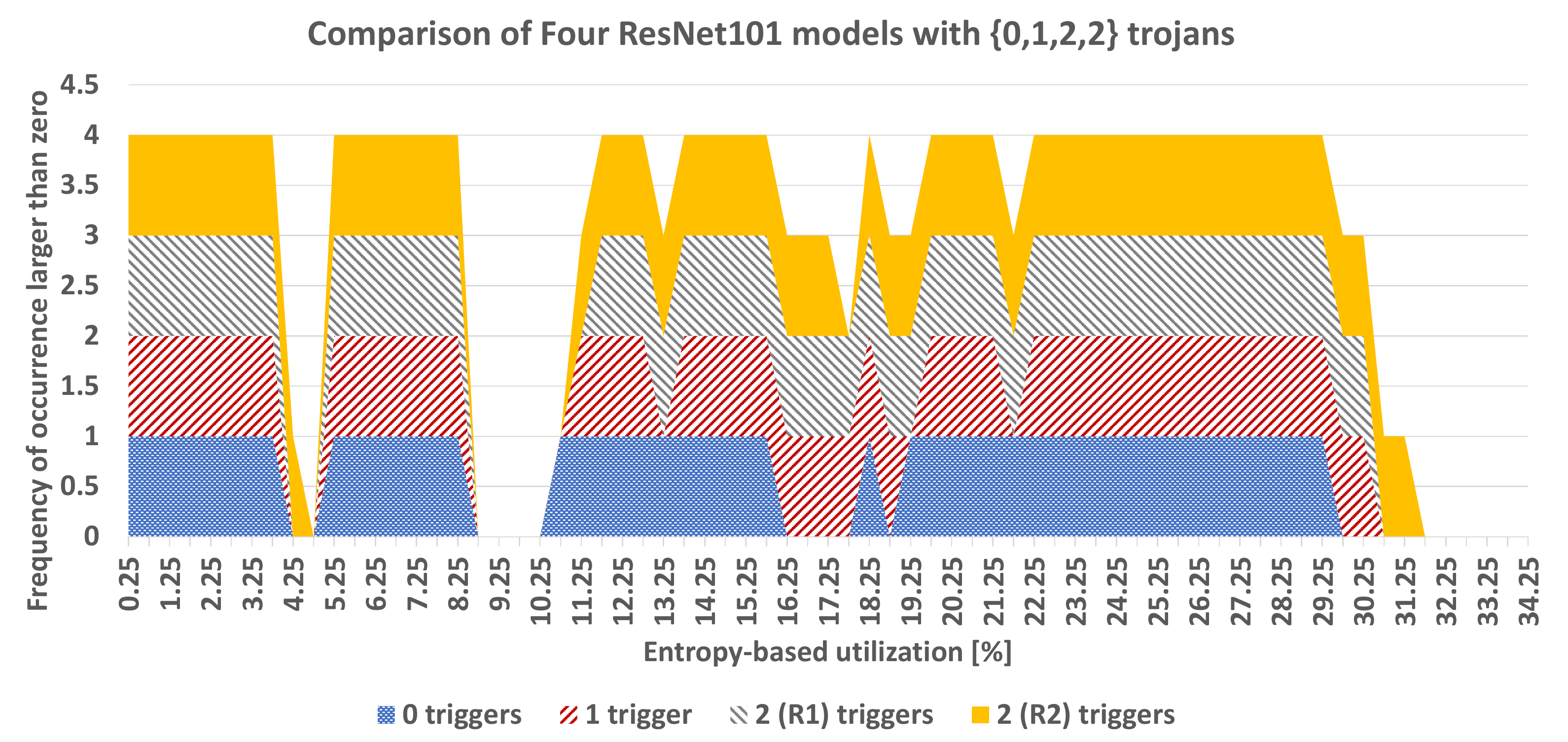}
  \centering
  \caption{Comparison of four ResNet101 models with zero, one, and two triggers (two replicates denoted as R1 and R2). }
  \label{fig:15b}
\end{figure}

Figure~\ref{fig:15a} shows a stacked histogram of an entropy-based utilization values for the four ResNet101 models. All four models have approximately the same distribution of utilization values over all encoded traffic classes. However, as can be seen in Figure~\ref{fig:15b}, there are utilization values in ranges $[16.0, 18.0] \cup [18.5,19.0]$ and $[29.5, 31.5]$ that are present in the poisoned models but are missing in the clean model. The utilization values in $[16.0, 18.0] \cup [18.5,19.0]$ are measured at the computation units labeled as maxpool, conv1, bn1, and ReLU (maximum pooling, convolution, batch normalization, and rectified linear unit). The utilization values in  $[29.5, 31.5]$ come from layer1.2.conv2  and layer1.2.bn2 in all poisoned models. In addition, the values in $[29.5, 31.5]$ are also measured in AI models poisoned with polygon triggers at the computation units labeled as layer1.1.conv2 and layer1.1.bn2.
Based on this granularity-level analysis, one can focus on the identified subset of computation units to explain the clean versus poisoned class encodings.





\subsubsection{Patterns detected in computation subgraphs:}

\begin{figure}
\includegraphics[
  width=12cm,
  height=6cm,
  keepaspectratio,
]{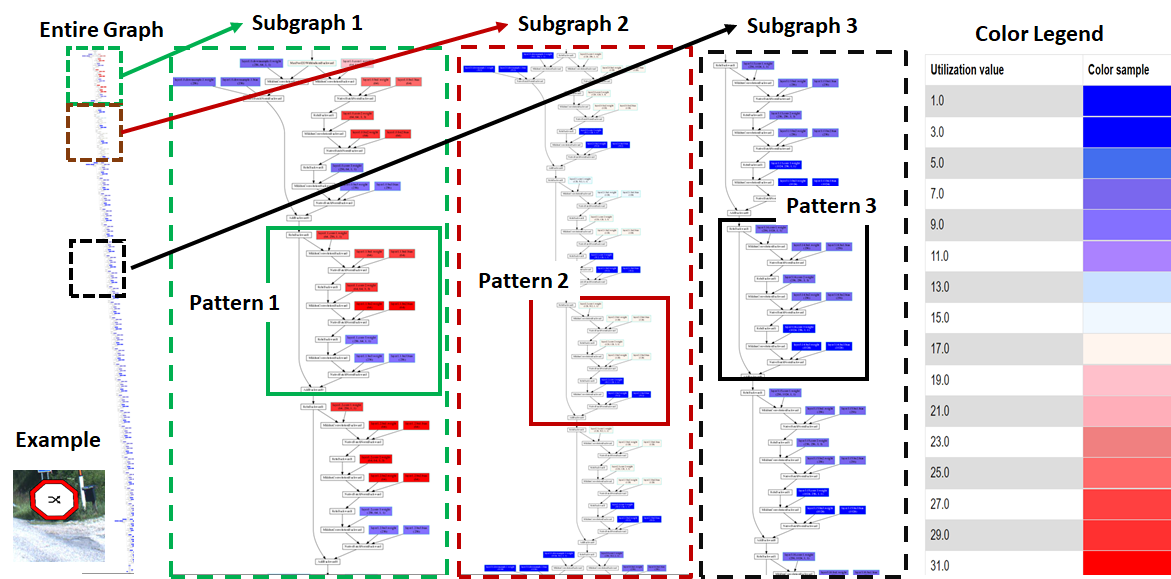}
  \centering
  \caption{Three distinct patterns of a class encoding in the clean trained ResNet101 model. An example traffic sign image of the encoded class is in the lower left. The three subgraphs are zoomed out sections of the entire graph to illustrate three repeating patterns of colors with the color legend on the right.}
  \label{fig:11}
\end{figure}

Class encodings in complex and large graphs can be rendered as vector or raster visualizations. Raster visualization faces a tradeoff between pixel resolution (graph node details) and display area (overall graph structure). After color-coding graph nodes based on utilization values, a raster image of ResNet101 architecture is around $2000 \times 20\,000$ pixels and a vector representation in Adobe PDF truncates the content due to the large dimension. Figure~\ref{fig:11} (left) shows an overview of the ResNet101 graph, which can be explored by zooming in and out as shown in the three zoomed out subgraphs (middle and right). In this class encoding of the 2500 traffic signs like the example shown in Figure~\ref{fig:11} (bottom left), one can identify three distinct subgraphs that repeat in the utilization-based color-coded ResNet101 graph.  We will focus primarily on Pattern 1 since it contains computational units (nodes) that are more utilized than those in Patterns 2 and 3.  All presented results in this subsection are derived from AI models predicting 75 traffic sign classes.  

In order to find utilization patterns, one is looking for repeating isomorphic subgraphs with the same utilization values assigned to all subgraph nodes. Such subgraphs encode class-specific parts in traffic sign classes, for example, several rotated curvatures in round shaped traffic signs using curve detectors \cite{olah2020zoom} in mono-semantic subgraphs or multiple spatial- and intensity-based traffic sign characteristics in poly-semantic subgraphs. We hypothesize that a presence of physically realizable trojans will perturb utilization patterns.  

\begin{figure}
\includegraphics[
  width=12cm,
  height=6cm,
  keepaspectratio,
]{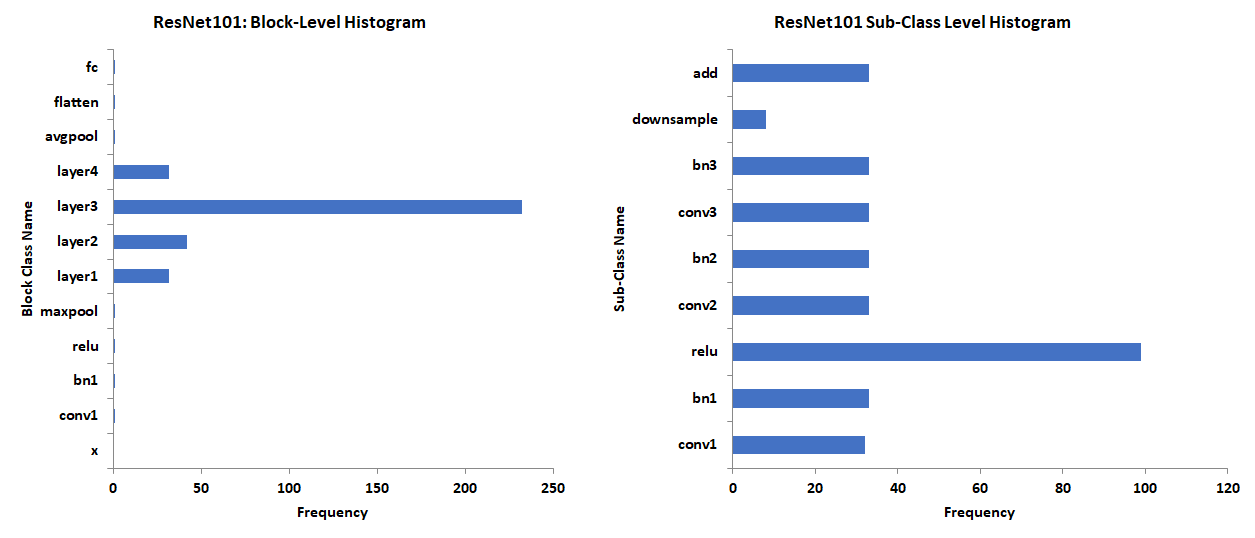}
  \centering
  \caption{Histograms of block class names and sub-class names of ResNet101 architecture modules}
  \label{fig:06}
\end{figure}

Automated finding of isomorphic subgraphs is a NP-complete problem. Thus, we used visualization as the main approach to identifying utilization patterns. The visualization approach is supported (a) by extracting a graph structure (sometimes) available in Torchvision representation while following the workflow for pseudo-coloring graph nodes shown in Figure~\ref{fig:04b} and (b) by identifying high-frequency graph nodes of the same computation unit type via histograms. First, if the Torchvision representation of a AI model computation graph uses block classes, then one can use all graph nodes (computational units) with a block class as a candidate subgraph for finding a pattern. Second, one can leverage automatically computed histograms of graph nodes for ranking the graph nodes based on their frequency as candidates in repeating subgraph patterns. 
Figure~\ref{fig:06} shows histograms of trace-based tree  computation units based on block class names (left) and sub-class names (right) of ResNet101 architecture.


\begin{figure}
\includegraphics[
  width=12cm,
  height=6cm,
  keepaspectratio,
]{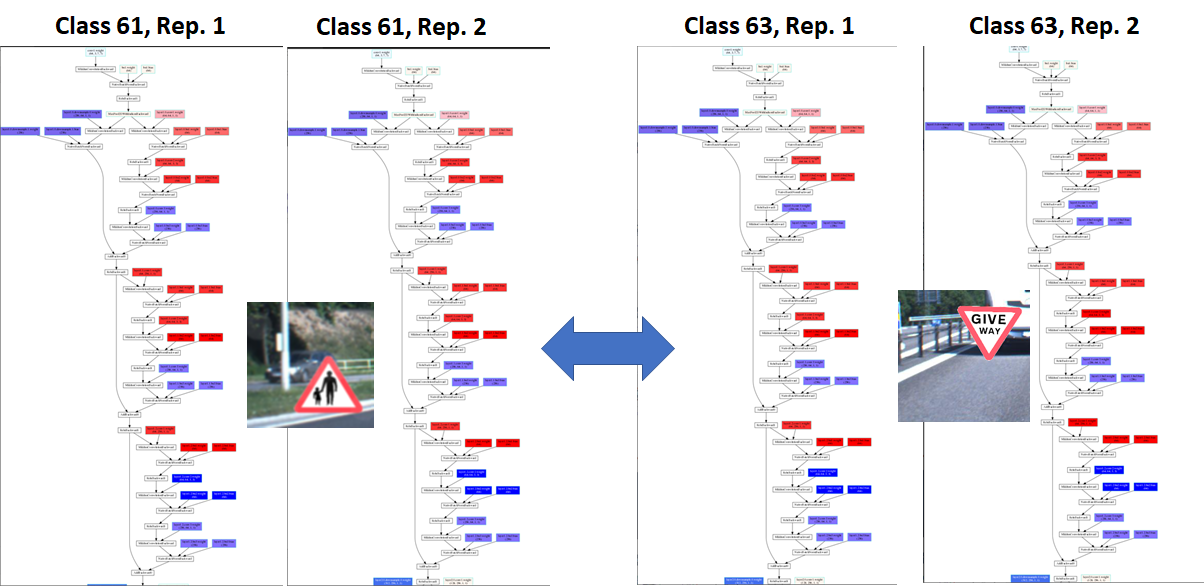}
  \centering
  \caption{Comparison of class encodings for two triangular traffic signs with constant color and triangular shape in two trained replicate AI models (ResNet101 architecture). The color legend is the same as in Figure~\ref{fig:11}.}
  \label{fig:12}
\end{figure}

\underline{Encodings of clean classes in computation subgraphs:} 
Figure~\ref{fig:12} shows invariant utilization patterns for (a) two clean classes $c=\{61, 63\}$ in the same trained AI model and (b) one class $c=63$ in two randomly initialized and trained AI models (Rep. 1 and Rep.2). The traffic signs are similar in triangular shape and red-white-black traffic colors, and dissimilar in the triangular orientation and black symbols inside triangles. The invariance of utilization patterns indicates that the uniqueness of each class encoding lies in the set of tensor-states while the utilization of computation units is constant (or the number of unique tensor-states per computation unit remains approximately constant in order to encode a class). 

\begin{figure}
\includegraphics[
  width=12cm,
  height=6cm,
  keepaspectratio,
]{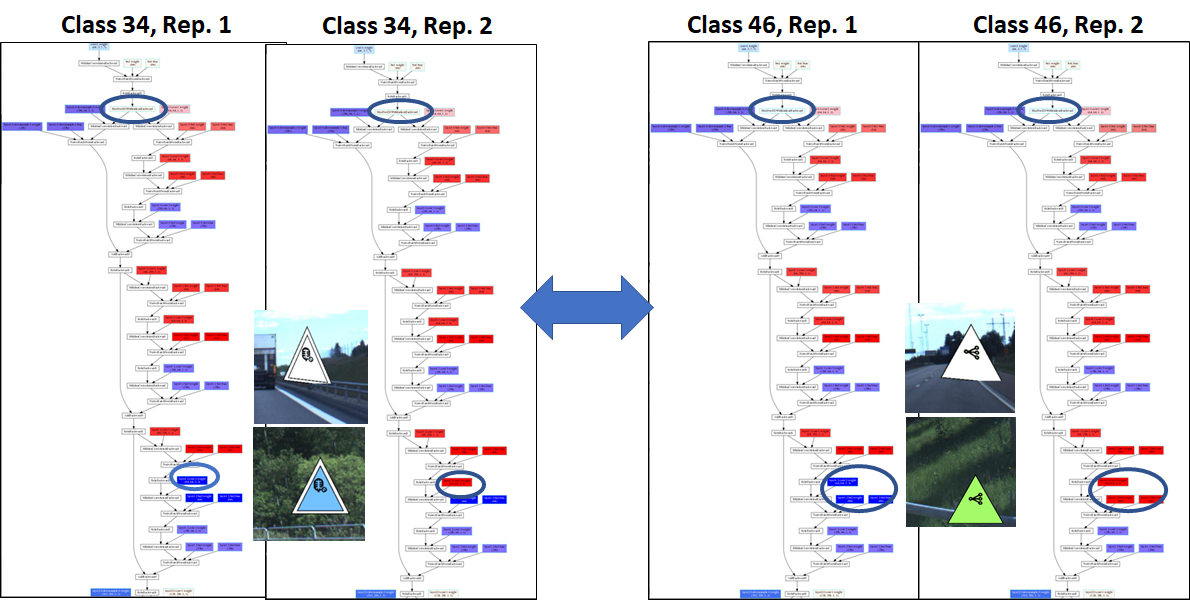}
  \centering
  \caption{Comparison of class encodings for two triangular traffic signs with varying color and constant triangular shape in two trained replicate AI models (ResNet101 architecture). The color legend is the same as in Figure~\ref{fig:11}.}
  \label{fig:13}
\end{figure}

On the other hand, when traffic signs that are characterized by fewer unique properties than in Figure~\ref{fig:12}, for example the signs shown in Figure~\ref{fig:13}, subsets of training images with common properties within a class will be encoded at various computation units to improve classification accuracy and robustness to a fewer unique characteristics. Such distributed class encodings will yield to varying number of unique tensor-states at each computation unit and hence varying utilization.  
Figure~\ref{fig:13} illustrates how varying color will cause perturbations of entropy-based utilization patterns in maxpool (circled at the top) and layer1.2.conv2, layer1.2.bn2.weight and layer1.2.bn2.bias (circled at the bottom) for two replicate retrained AI models (randomly initialized) and two similar traffic signs. The two sample training images highlight only the color variability and the presence or absence of the white rim (left vs right).

\begin{figure}
\includegraphics[
  width=12cm,
  height=6cm,
  keepaspectratio,
]{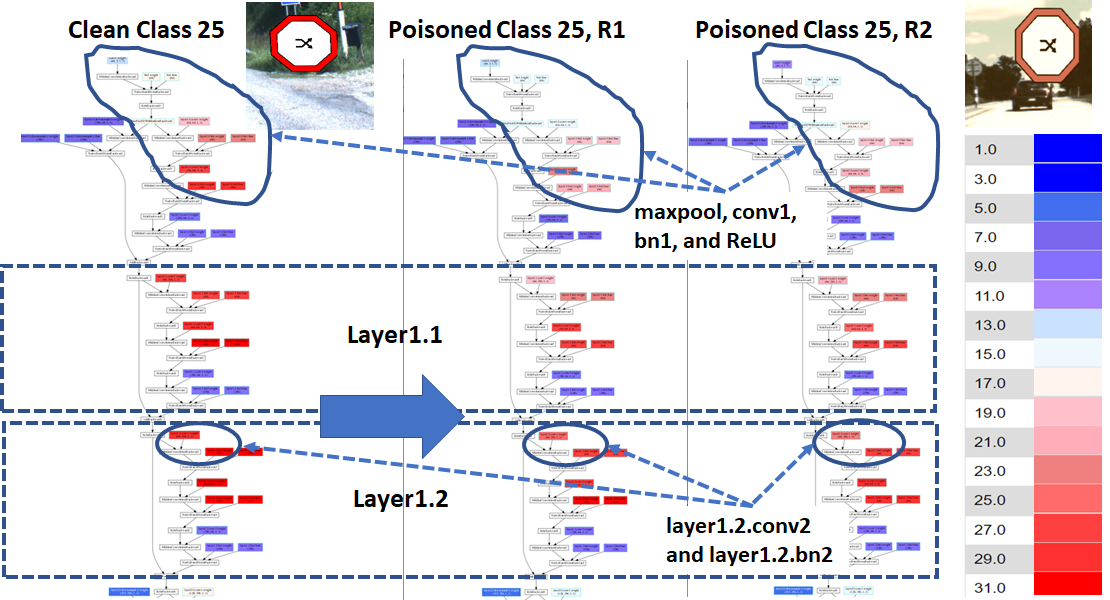}
  \centering
  \caption{Comparison of a clean class encoding evaluated with clean images (left) and of poisoned class encodings in two trained replicate AI models (ResNet101 architecture) evaluated with poisoned images (middle and right). The circles show the variability of utilization in the initial graph nodes and layer1.0 in two poisoned class encodings. The rectangles show the utilization pattern change between clean and poisoned class encodings.}
  \label{fig:14}
\end{figure}

\underline{Encodings of clean versus poisoned classes in computation subgraphs:} 
Figure~\ref{fig:14} shows the comparison of clean class encoding $c=25$ (left) and two replicate class encodings of $c=25$ with Kelvin Instagram filter as a trigger (middle and right) in the ResNet101 architecture.  
Based on the AI model fingerprint analyses in Section~\ref{pattern:graph}, Instagram filters and polygons as triggers present themselves in the initial maxpool, conv1, bn1, and ReLU computation units. Varying utilization (different from the clean class encoding) can be observed in Figure~\ref{fig:14} with the circles enclosing maxpool2d, ReLU, conv1.weight,
layer1.0.conv1.weight,
layer1.0.conv2.weight,
layer1.0.bn2.weight, and 
layer1.0.bn2.bias. The color coding goes from dark blue to dark red or from 1$\,\%$ to 31$\,\%$ of entropy-based utilization (see the color legend in Figure~\ref{fig:11}).

Regarding the subgraph pattern 1 shown in Figure~\ref{fig:11}, the trigger of Kelvin Instagram filter type breaks the pattern between layer1.1 and layer1.2 as highlighted with two dash-line rectangles in Figure~\ref{fig:14}. Since the Kelvin Instagram filter reduces the color spectrum to the earth tones of green, brown, and orange, this will reduce the number of unique tensor-states and, hence, reduce the utilization of some computation units.


It should be noted that patterns in subgraphs could also be studied for groups of traffic sign classes. For instance, one can group traffic signs by shapes to identify subgraphs that serve the same purpose but are parametrized differently. This type of pattern analyses must overcome the memory challenges as a larger number of classes implies a larger number of training images and an even larger number of unique tensor-states to keep track of. Based on our experiments, we could evaluate up to four classes together or $10\,000$ training images of size $256 \times 256$ pixels, and their corresponding unique tensor-states. 

%

\subsubsection{Patterns detected in computation units:}

\begin{figure}
\includegraphics[
  width=12cm,
  height=6cm,
  keepaspectratio,
]{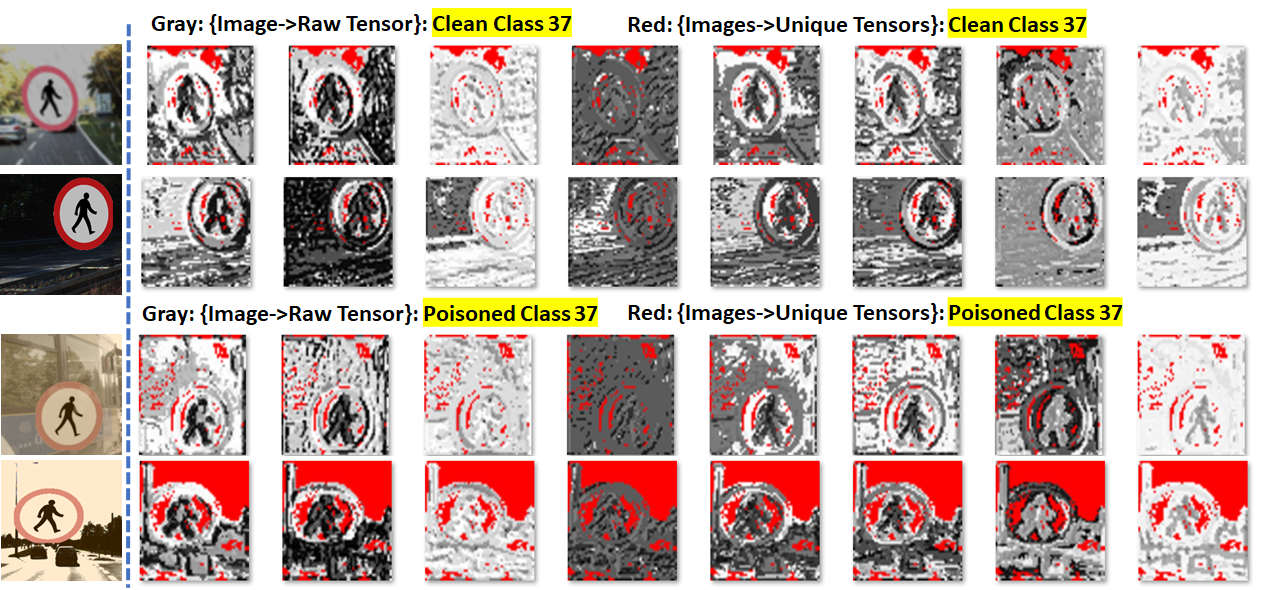}
  \centering
  \caption{Visualization of tensor-state values in red for two sample clean (top two rows) and poisoned (bottom two rows) images from the same class in layer1.2.conv2 of ResNet101 that occur more than 100 times in 2500 clean images (top two rows) or in 2500 poisoned images (bottom two rows). 
  }
  \label{fig:16a}
\end{figure}

\begin{figure}
\includegraphics[
  width=12cm,
  height=6cm,
  keepaspectratio,
]{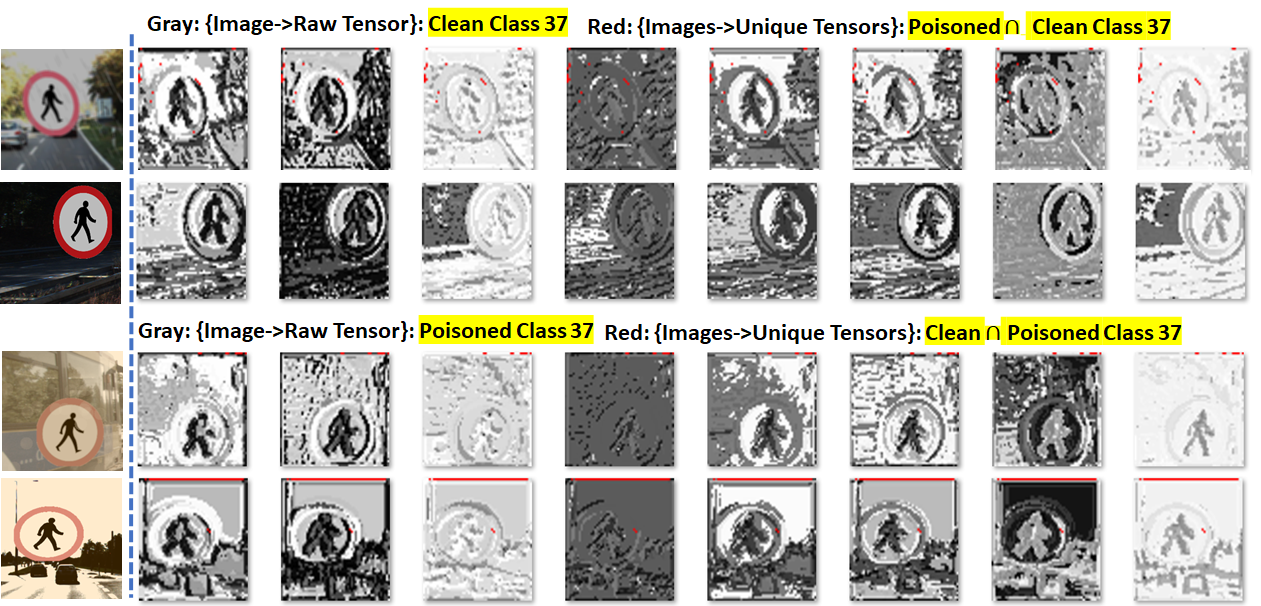}
  \centering
  \caption{Visualization of tensor-state values in red for two sample clean (top two rows) and poisoned (bottom two rows) images from the same class (\emph{STOP pedestrian crossing road} signs) in layer1.2.conv2 of ResNet101 that occur more than 100 times in both 2500 clean and 2500 poisoned images. 
  }
  \label{fig:16b}
\end{figure}

Similar to Figures~\ref{fig:ts01} and \ref{fig:ts01b}, we compared the tensor-states characterizing clean and poisoned classes in  Figures~\ref{fig:16a} and \ref{fig:16b}. The comparison of clean and poisoned classes is shown for  the same \emph{STOP pedestrian crossing road} sign with or without applied Kelvin Instagram filter as a trigger.  Figure~\ref{fig:16a} (top two rows) illustrates that the common tensor-state values within a clean class  correspond to sky, parts of a road without shadows, and several pixel clusters inside the traffic sign.  After applying the Kelvin Instagram filter, Figure~\ref{fig:16a} (bottom two rows), the common tensor-state values within a poisoned class are dominated by the earth tones of green, brown, and orange, which leads to a merger of semantically distinct regions, such as sky, parts of the road, and interior of the traffic sign. The image in the lowest row of Figure~\ref{fig:16a} has a significant number of red pixels suggesting that it consists of many features common across all poisoned images.

The objective of Figure~\ref{fig:16b} is to visualize with red pixels any common tensor-states across clean and poisoned classes.  All four rows in Figure~\ref{fig:16b} show almost no red pixels except from a few pixels from the yellowish tree and from a red rim of the traffic sign in the top row left image. 
Since the Kelvin Instagram filter affects every pixel in a training image, the overlap of high-frequency tensor-state values between clean and poisoned images is only 35 tensor-state values and almost none in the area of the \emph{STOP pedestrian crossing} traffic sign. In other words, although perceptually the areas of clean and poisoned traffic signs are very similar, the features characterizing each class as generated by the computation unit layer1.2.conv2.weight are completely different. Furthermore, since the Kevin Instagram filter blurs pixel values but makes their color more similar to each other, there are less unique tensor-state values in poisoned images than in clean images due to blur but more tensor-state value with high values due to color similarities.

\begin{figure}
\includegraphics[
  width=12cm,
  height=6cm,
  keepaspectratio,
]{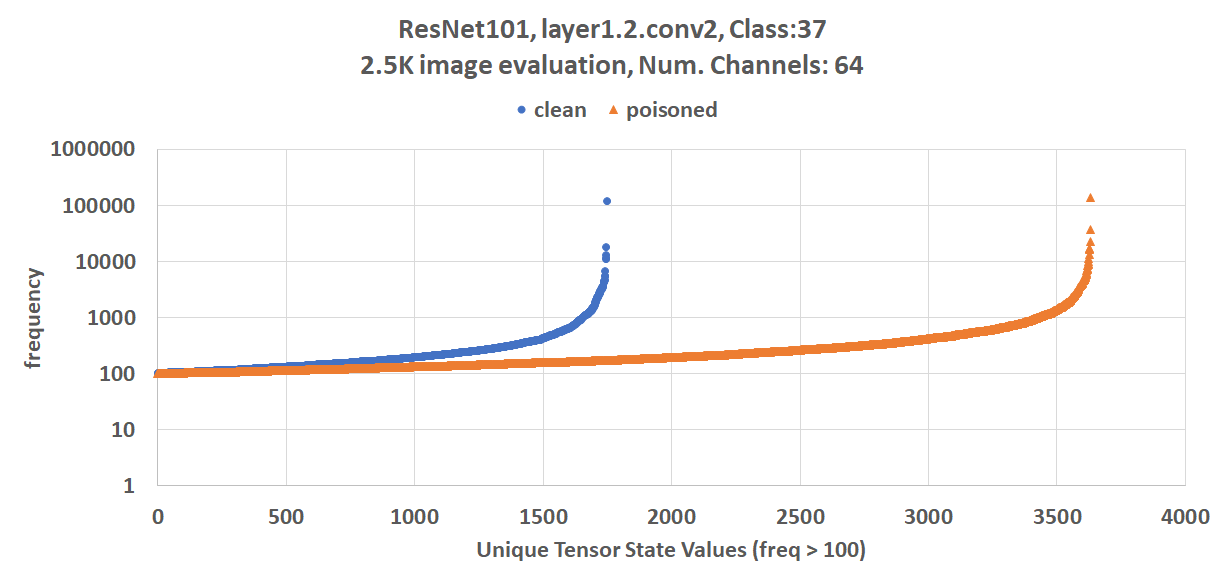}
  \centering
  \caption{Histograms of unique tensor-state values in layer1.2.con2 of ResNet101 that occur more than 100 times in 2500 clean images (blue circles) and in 2500 poisoned images (orange triangles). Unique tensor-states are sorted by the frequency but are not matched across clean and poisoned images. Example images are shown in Figure~\ref{fig:16a}.}
  \label{fig:16c}
\end{figure}

Another approach to inspecting the class encodings is via histograms of tensor-state values. Table~\ref{table:05} summarizes statistics of the number of unique tensor-state values in layer1.2.conv2 for clean and poisoned training images in poisoned ResNet101 AI models. To scale down the visualization requirements on a histogram with $5.4$ million bins, we threshold the bins based on the tensor-state value counts. Figure~\ref{fig:16c} shows the histogram visualization for the threshold value equal to $100$ using Microsoft Excel. The frequency (count) along a vertical axis is shown on a logarithmic scale to accommodate the wide range of values. The horizontal values are sorted by frequencies. 

The tabular and histogram visualizations allow us to observe the differences in the number of tensor-state values between clean and poisoned images as a function of unique tensor-state frequencies. The utilization values reflect the differences in the numbers of unique states and the distributions of their frequencies. While clean images give rise to more unique tensor-states than poisoned images $(5 \, 366 \, 576 - 3 \, 734 \, 889 = 1 \, 631 \, 687)$, they are also characterized by lower frequencies of  tensor-states as shown in Table~\ref{table:05}. We observe in Figure~\ref{fig:16c} that both distributions of sorted clean and poisoned tensor-states follow the same trend and hence the number of unique tensor-states becomes the key factor for a utilization value. This explains why the encodings of clean images yield a higher utilization  $\eta_{j=layer1.2.conv2}$ in the ResNet101 AI model than the encodings of poisoned images (see Table~\ref{table:04}, last row).
  
\begin{table}
  \caption{Number of unique tensor-state values invoked by 2500 clean images and 2500 poisoned images with frequencies higher than 0, 1, 10, and 100 in layer1.2.conv2 of ResNet101 (image examples are shown in Figure~\ref{fig:16a}).}
  \label{table:05}
\centering
\begin{tabular}{|c | c | c |}
\hline
Num. unique state values 	& Clean &	Poisoned   \\
\hline
\makecell{$>0$} & \makecell{$5 \, 366 \, 576$} & \makecell{$3 \, 734 \, 889$}  \\
\hline
\makecell{$>1$} & \makecell{$531 \, 183$} & \makecell{$645 \, 297$} \\
\hline
\makecell{$>10$} & \makecell{$35 \, 047$} & \makecell{$57 \, 985$} \\
\hline
\makecell{$>100$} &  \makecell{$1750$} & \makecell{$3633$}  \\
\hline
\makecell{$\eta_{j=layer1.2.conv2}^{entropy}$} & \makecell{$33$} & \makecell{$30.6$}  \\
\hline
\end{tabular}
\end{table}

\section{Discussion}
\label{discussion}

\underline{Trojan injections and detections:}
The examples shown in Figures~\ref{fig:16b} and \ref{fig:16c} have some implications on trojan injections and detections. Two classes of images (clean and poisoned) with identical semantic labels according to a human visual inspection are encoded with completely different tensor-state representations. This evidence is presented for one of many Instagram filters as a trigger and at one of many computation units in one of many AI model architectures. While we need additional studies to generate more class encoding examples and their dependencies on trigger types, we hypothesize that such relationships between training data and AI model can improve robustness of trojan injections and detections.    

Assuming that trojans can be defined in a very large space of spatial and color image configurations, one can search for pairs of visually insignificant (or imperceptible) image changes and non-overlapping class encodings in order to improve robustness and convergence of trojan injections. Finding non-overlapping class encodings is related to the choice of AI model architecture (i.e., the choice of computation units and their connectivity) and can be framed as a type of a network architecture search (NAS) problem. If image classification training data and AI model architectures are understood in terms of utilization-based class encodings, then trojans can be designed in the image space to be encoded in underutilized computation units during training. 
On the other hand, trojan detectors can become more efficient in predicting poisoned AI models by first identifying computation units of interest, such as with pruning approaches \cite{Liu2018a}, \cite{BabakHassibi1992}, \cite{Li2017}, \cite{Bajcsy2021a}. Trojan detectors can then be specialized by investigating subgraphs that (a) encode classes of trojans and (b) make the trojan search space for any detector much smaller. 

\underline{Understanding overlapping characteristics of class encodings:}
We have studied overlapping characteristics of two clean classes or clean and poisoned classes. There are additional challenges in exploring overlapping characteristics of multiple encoded classes due to the much more complex Venn diagram for an AI model predicting a number of classes (equal to $C$). The challenges are not only in a larger number of visualizations, but they are also in the large cost of matching states across all combinations of classes and their lists of unique tensor-states at the order of several millions (see Table~\ref{table:05}). Furthermore, given the cost of generating ``explainable'' class encodings and its purpose of delivering trusted predictions, one may be limited by the size of training data and complexity of AI models in the future. While explanations of class predictions have a ceiling on information content \cite{Sarkar2022}, the AI models do not have a ceiling on complexity. In other words, explanations must compress AI model complexity at a higher rate with the increasing model complexity \cite{Sarkar2022}. Therefore, one would select only a few AI models with datasets to be fully characterized and trusted.

\underline{Natural trojans:}
We hypothesize that the tensor-states for clean and poisoned classes can explain why trojan detectors can be misled by natural trojans. Natural trojans in AI models are informally defined as those that yield a high false positive rate in predicting poisoned AI models. If one traffic sign class is trained on images characterized by two distinct groups, for instance, red and blue color traffic signs, then (a) trojan detectors might incorrectly rely on a correlation between the presence of trojans and a multi-modal distribution of training data for a single class, and (b) trojans can be injected by relabeling output labels for one group and retraining the model.

\underline{Detectability of trojans in poisoned AI models:}
While we showed a minimum tensor-state overlap between clean and poisoned class encodings in Figure~\ref{fig:16b} for semantically equivalent traffic signs of \emph{STOP pedestrian crossing} sign, it is not clear whether detecting encodings of hidden classes is computationally feasible without a priori knowledge about types of triggers, types of foreground, and types of background by the trojan detector designers.  One can view  trojan detection as an outlier detection problem \cite{Dietterich2022} and aim at establishing probably approximately correct (PAC) style learning bounds on the outlier detection under general assumptions. However, such limited guarantees on a successful trojan detection pose vulnerability risks as reported by Sun et al. \cite{Sun2020} by demonstrating a patch attack in backdoored ``broken'' classifiers. Furthermore, Goldwasser et al. \cite{Goldwasser2022} has shown two frameworks of planting undetectable backdoors with incomparable guarantees (i.e., finding a trojan cannot be solved in polynomial time) for a class of ReLU networks. This topic of trojan detectability remains an open research problem.

\underline{Evaluation data for better explainability:}
Table~\ref{table:03} shows three types of evaluation data. Our evaluation sets did not contain pairs of clean and poisoned images that would differ only by the trigger. In other words, the background image and the fusion parameters were always different. One could understand the clean versus poisoned class encodings even better if the sets had matching pairs of clean and poisoned images. It would also be possible to explore the patterns in poisoned AI models with images from clean classes and with injected triggers. 

\underline{Evaluation of AI model explainability:}
While we have shown three methods for explainable AI at graph, subgraph, and tensor-state granularity levels, we have not done quality evaluations of AI model explainability via model parameter and data randomization tests  \cite{Adebayo2018}. It remains to be demonstrated how illustrated tensor-states, utilization patterns in subgraphs, and utilization distributions in multiclass prediction models actually reflect a true explanation of the traffic sign prediction.

\underline{Estimation of maximum number of classes that can be encoded:}
Given the number of training image per class and the tensor dimensions of a computation unit $v_{j}$, it is possible to compute the maximum number of classes that can be uniquely encoded by a graph unit $v_{j}$. This was  calculated for the example tensor-state $(1,64,56,56)$ to be $C_{layer1.2.conv2}^{MAX}=\frac{2^{64}}{56*56*2500}  \approx 2.35*10^{12}$ (a terascale count of traffic sign classes). This number is purely theoretical since  $2500$ images of one type of a traffic sign would not fully represent the entire spectrum of imaged traffic signs in practice. In addition, it is not clear how the graph connectivity would be incorporated into the estimation.  

\underline{Computational constraints on utilization measurements:}
Equations~\ref{eq:04} and \ref{eq:05} summarize the formulas for evaluating computational challenges. 
For example, the computational cost of inferencing $M=2500$ images with $n(a)=286$ probes in $a=$ResNet101 takes on average 24.46 minutes ($\widehat{T}(F_{a}(\vec{x}_{i})=0.587)$s
while the memory consumption can reach up to 140.6 GB for one model from the TrojAI Challenge \cite{TrojAI2022} (AI models in Round 4).

In this case, the input image size $\vec{x}_{i}$ was used to approximate output tensor size $\max_{j}(D_{j}^{Out})$ to be $256 \times 256 \times 3=196\,608$ Bytes (Sample images are cropped to $224 \times 224 \times 3$ in Round 4). The average time estimate was obtained experimentally on CPU: AMD Ryzen Threadripper 3970X 32-Core Processor; MemTotal: $264$ GB
and GPU: NVIDIA GeForce; MemTotal: $246$ GiB. 
Measuring utilizations requires a processing time proportional to 1008 AI trained models, hundreds of  computation units in each of the 16 included AI architectures, 2500 images per class, and between 15 and 45 predicted traffic sign image classes.
These numerical values indicate our current hardware is limited to handling about 10,000 tensor-states in memory for $a=$ResNet101 (Virtual Memory$=562.3$ GB according to Equation~\ref{eq:05}). 


\section{Summary}
\label{summary}
We have introduced the concept of AI model utilization for the purpose of (a) delivering explainable AI models at graph, subgraph, and tensor-state granularity levels, and (c) accelerating injection and detection of poisoned AI models. We defined a mathematical framework for computing three deterministic and statistical AI model utilization metrics. Appendix~\ref{reasoning} describes the relationship of these metrics to other theoretical approaches that have described AI models. Furthermore, we implemented a suite of tools for measuring utilizations of each computation unit in a computation graph and visualized the utilization measurements as matrices (AI model fingerprints), color-coded graphs, and a sequence of images representing a multidimensional array. 

Specifically, we explained the utilization-based class encodings for clean and poisoned classes from the TrojAI Challenge (Rounds 1-4) \cite{IARPA2020}. We concluded that while clean and poisoned images can clearly be  classified into the same semantic traffic sign category, a poisoned AI model would have completely independent tensor-states for clean versus poisoned traffic sign images (see Figure~\ref{fig:16a} versus Figure~\ref{fig:16b}).  In addition, the tensor-state values observed as common to all images defining a class come from foreground and background image regions. Note that the importance of both types of regions for accurate prediction has been reported by Xiao et al. cite{Xiao2020}. The utilization-based subgraphs for clean and poisoned classes illustrated that the utilization patterns changed (see Figure~\ref{fig:14}) and could become a coarse indicator of tampering with training data and distributing a poisoned AI model. Similarly, presence or absence of utilization values in all class encodings represented by an AI model (i.e., AI model fingerprint) allowed us to focus on specific subgraphs and hence reduce the search space for explaining differences between clean and poisoned classes. Finally, we documented several cost versus explainability tradeoffs in terms of computational requirements. As there is a race between explainability costs and the growth of training datasets and AI model sizes \cite{Sarkar2022}, we found the path toward explainability only if training images, inferenced images, and trained AI models are available for in-depth analyses. If other than known training and inferenced images are presented, then trust in predictions would be limited due to a tug of war between trojan detectors and trojan detectability (see Discussion section).

In the future work, we will analyze existing AI models that have been identified to contain natural triggers. We also plan to document the characteristics of physically realizable triggers found in traffic sign classifications models. Our future work is motivated by delivering fully trusted AI models for life-critical applications. In the context of our presented work, fully trusted AI models imply access to all tensor-states (activation maps) at each graph node and for each training image, as well as their overlaps with tensor-states invoked by other predicted classes (i.e., tensor-states common to two and more classes). Furthermore, we plan to contribute to symbolic representations of graphs and subgraphs in the context of training images consisting of semantically meaningful objects. 

\section*{CRediT authorship contribution statement}
Peter Bajcsy: Conceptualization, Theory, Methodology, Software, Experiments, Data analyses, Writing - original draft preparation; 
Antonio Cardone: Data analyses - hypothesis testing, Writing - reviewing;
Chenyi Ling: Data analyses - hypothesis testing;
Philippe Dessauw: Data - AI model training, Writing - reviewing;
Michael Majurski: Data - AI model training, Writing - reviewing; 
Tim Blattner: Data - AI model training; 
Derek Juba: Hardware - AI model training;
Walid Keyrouz: Writing - reviewing and editing.

\section*{Declaration of competing interest}
The authors declare that they have no known competing financial interests or personal relationships that could have appeared to influence the work reported in this paper.

\section*{Acknowledgement}
The funding for all authors was provided by the Intelligence Advanced Research Projects Activity (IARPA): IARPA-20001-D2020-2007180011. We would like to acknowledge the contributions of Mylene Simon and Ivy Liang to develop an interactive web application for online comparison of AI model fingerprints. We would also like acknowledge Peter Fontana and Joe Chalfoun from National Institute of Standards and Technology for providing additional comments on the manuscript.

\section*{Disclaimer}
Commercial products are identified in this document in order to specify the experimental procedure adequately.
Such identification is not intended to imply recommendation or endorsement by the National Institute
of Standards and Technology, nor is it intended to imply that the products identified are necessarily the best available for the purpose.

\bibliographystyle{unsrt}
\bibliography{nn_utilization}

%
%


\appendix

\section{Theoretical Reasoning and Relationships to Past Work}
\label{reasoning}
\underline{Binary tensor-states in AI computation graphs:} 
While one  computation unit in an AI computation graph produces real-value numbers, they can be binarized by thresholding at zero during measurement extraction. The thresholding operation reduces the real-value variability to $\{0, 1\}$ by discretizing the output real values. 
One can model the discretization step according to Hopfield Neural Networks (HNN) \cite{Hopfield1982} where a  computation unit is viewed as an associative content-addressable memory for storing information bits about training data points \cite{Hopfield1985}. The HNN model was motivated by collective properties of neurons in neurobiology adapted to integrated circuits \cite{Hopfield1982}, and the more recent work on dense associative memory \cite{Krotov2016} and Hopfield layers \cite{ramsauer2021hopfield}. The thresholding value at zero is motivated by zero-centered normalizations of input data and the majority of nonlinear activation functions being centered around zero (i.e., tanh, sigmoid, rectified linear unit (ReLU), or leaky ReLU). 

\underline{Utilization With Respect to State Representation Power of a  Computation Unit:}
A particular  computation unit in an AI computation graph can represent a maximum number of states depending on the number of outputs. For example, a fully connected layer (one  computation unit) consisting of two nodes (two scalar outputs) can represent no more than four states, such as $\{00, 01, 10, 11\}$.  This fully connected layer would be fully utilized if all four states were used for predicting output labels. 

In order to define the utilization of a  computation unit mathematically, one can use the parallels drawn by Bajcsy et al. \cite{bajcsy2021} between neural network and communication fields in terms of (a)  computation unit maximum representation power and channel capacity in communications and (b)  computation unit utilization and channel efficiency while leveraging the universal approximation theorem \cite{Hornik1991} and the source coding theorem \cite{Shannon1948}.
Using the theorems, a computation unit $v_{j}$ with $D_{j}^{Out}$ nodes has the maximum representation power (or  computation unit capacity) of  $n_{j} = 2^{D_{j}^{Out}}$ possible states.

The utilization definition can also be related to Vapnik-Chervonenkis (VC) dimension of a class of binary classifiers $H$ from the image space $\chi$ to the label space $\{0,1\}$;  $H: \chi \rightarrow \{0,1\}$   \cite{Blumer1989}, \cite{Bartlett2017}. The VC dimension $VCdim(H)$ is defined as the size $m$ of the largest shattered set of inputs $\{\vec{x}_{1},..., \vec{x}_{m}\}$ such that the growth function 
$\Pi_{H}(m)= \max_{\vec{x}_{1},..., \vec{x}_{m} \in \chi} | \{ (h(\vec{x}_{1}),..., h(\vec{x}_{m}) ): h \in H \} |= 2^{m}$  \cite{Bartlett2017}.  
For real-valued functions $\mathcal{F}$ present in neural networks, one can define $VCdim(\mathcal{F}) = VCdim(\{sgn(f):f \in \mathcal{F}\}$, which is achieved by thresholding at zero in the utilization computation. The pseudodimension of $\mathcal{F}$ or $Pdim(\mathcal{F})$ has been proven to satisfy $VCdim(\mathcal{F}) \leq Pdim(\mathcal{F})$ \cite{Anthony2009}.
For example, a convolutional computation unit $v_{j}$ with the output tensor $(Channels, Width, Height)$ per input image $\vec{x_{i}}$ would yield $VCdim(\mathcal{F}(v_{j}))=Channels$ after binarizing the output values. The $VCdim$ values are used to quantify utilization of each computation unit as documented in Section~\ref{methods}.  Our work does not currently leverage the possibility of using lower and upper bounds on $VCdim(\mathcal{F})$ for an entire neural network. These bounds have been derived for AI architectures with computation units characterized by constant and known non-linearities (piece-wise constant, piece-wise linear, and piece-wise polynomial \cite{Bartlett2017}) with respect to the number of layers $L$ and parameters $W$. While in practice there is no guarantee that  computation units in one architecture are not characterized by a mixture of non-linearity types, it would be possible to verify the theoretical bounds for a combination of $L$ and $W$ in a variety of existing popular AI architectures.

\underline{Analytical and Statistical Utilization Definitions:}
Given a set of training data points activating a computation unit in an AI computation graph, the  computation unit will generate unique output states at varying frequencies. Considering that training datasets are formed by sampling and split into train, validation, and test subsets, the utilization value would vary depending on the sampling techniques and subset of samples. Thus, one can define the utilization $\eta_{j}$ of a  computation unit $v_{j}$ using analytical or statistical views since the states generated as outputs of a  computation unit by evaluating a set of training data points can be interpreted as deterministic states or as a statistical distribution of states.

An analytical definition of  computation unit utilization $\eta^{state}$ is directly derived as a ratio of a number of unique tensor-states and a computation unit capacity. One could aim at optimizing a network architecture search (NAS) \cite{nasbench2019} such that every  computation unit reaches maximum utilization $\eta^{state}=1.0$ over all training data points and all predicted labels. 
A statistical definition of  computation unit utilization is derived from a tensor-state distribution shape and aims at maximum normalized entropy of the state distribution for each predicted label (entropy-based utilization) $\eta^{entropy}$ or at minimum Kullback–Leibler (KL) divergence of the state distribution from a uniform distribution allocated per predicted label $\eta^{KLDiv}$. The use of Entropy and KL divergence~\cite{Kullback2017} are borrowed from the source coding theorem~\cite{Shannon1948}.  

\underline{Ordering Utilization Values in Class Encodings and AI Model Fingerprints:}
 For image classification AI models, a training dataset consists of one set of input images per output class label. Performing an inference on a set of training images per class label will produce one of the three utilization values per measurement probe (e.g., right after each  computation unit of a computation graph). The order of probes can be determined based on the flow of input images through a computation graph since it is deterministic and well-defined during an inference computation. Given a sorted (ordered) list of probes and their utilization values, \emph{a utilization-based class encoding} is uniquely defined.
 
An AI model predicts multiple class labels, and, therefore, utilization evaluations of the same AI model will consists of multiple class encodings. Although class labels have semantic annotations (i.e., cat or dog), they are typically numerically labeled as well, and, therefore, they can be ordered. Given a sorted (ordered) list of classes by numerical labels, \emph{a utilization-based fingerprint of an AI model} is well-defined as a matrix of ordered utilization-based class encodings.
 

\section{Verification of Utilization Measurement Properties}
\label{properties}

\begin{figure}
\includegraphics[
  width=12cm,
  height=6cm,
  keepaspectratio,
]{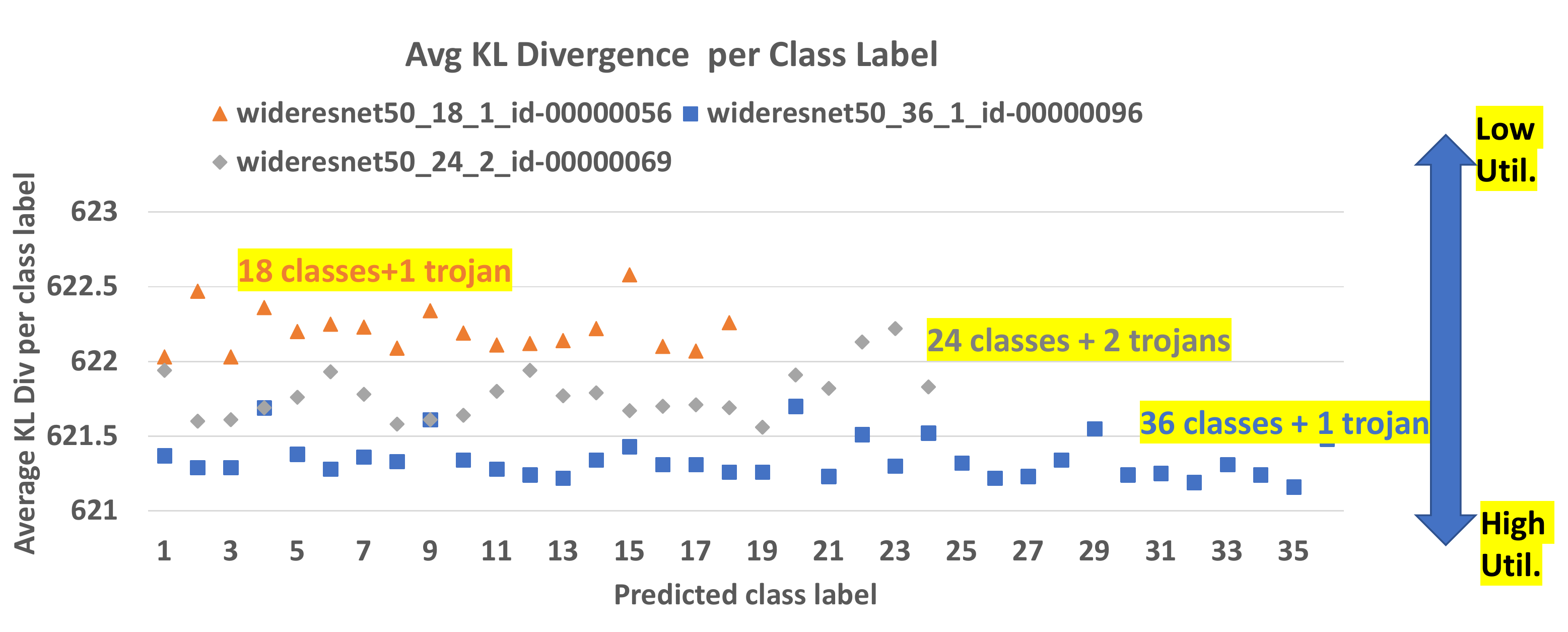}
  \centering
  \caption{Average KL Divergence utilization over all probes as a function of the number of predicted classes in WideResNet50 architecture.}
  \label{fig:08}
\end{figure}

\begin{figure}
\includegraphics[
  width=12cm,
  height=6cm,
  keepaspectratio,
]{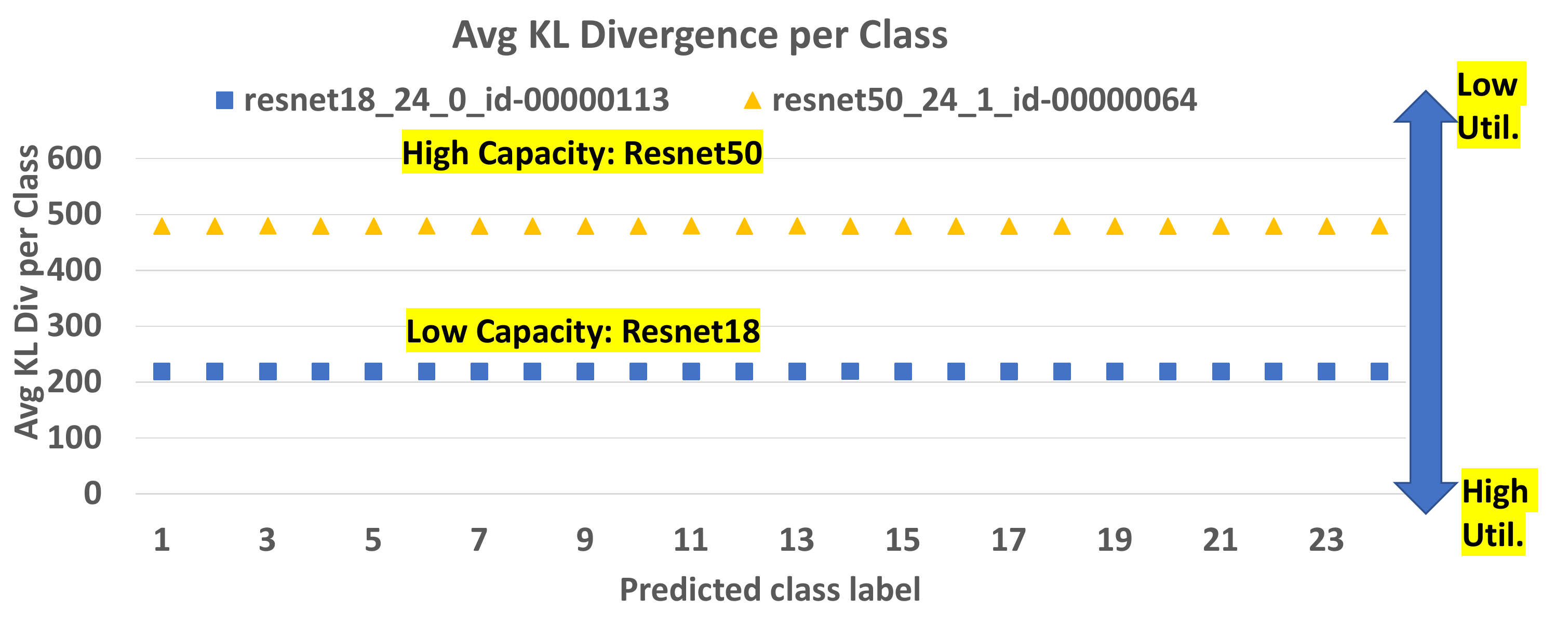}
  \centering
  \caption{Average KL Divergence utilization over all probes as a function of ResNet AI model architectures. ResNet18 and ResNet50 differ by including 20 or 50 convolutional layers affecting their modeling capacity.}
  \label{fig:09}
\end{figure}

Utilization measurements must satisfy a few expected properties listed below:
\begin{enumerate}
\item \emph{Inferenced data:} Average utilization must be nondecreasing with increasing number of data points used for utilization measurements - see Figure~\ref{fig:07}.
\item \emph{Predicted classes:} Average utilization must be nondecreasing with increasing number of predicted classes - see Figure~\ref{fig:08}.
\item \emph{AI model capacity:}Average utilization must be nondecreasing with decreasing capacity of an AI model - see Figure~\ref{fig:09}.
\end{enumerate}
Since the KL Divergence-based utilization measures `non-utilization' (or inefficiency), the trends demonstrating the properties are reversed. The low and high utilization values are highlighted along a vertical axis on the right side of Figures~\ref{fig:08}, \ref{fig:09},  and \ref{fig:07}.

\section{Identification and Placement of Utilization Measurement Probes}
\label{probe_placement}

Every utilization measurement (every probe) has its associated cost in terms of computer memory and execution time. The placement of measurement probes can be optimized so that the number of probes is minimal and the explainability is maximal. The challenges in optimal placement come from the fact that AI model computation graphs are not only structurally very complex but also very heterogeneous in terms of  computation units. 

\begin{figure}
\includegraphics[
  width=12cm,
  height=6cm,
  keepaspectratio,
]{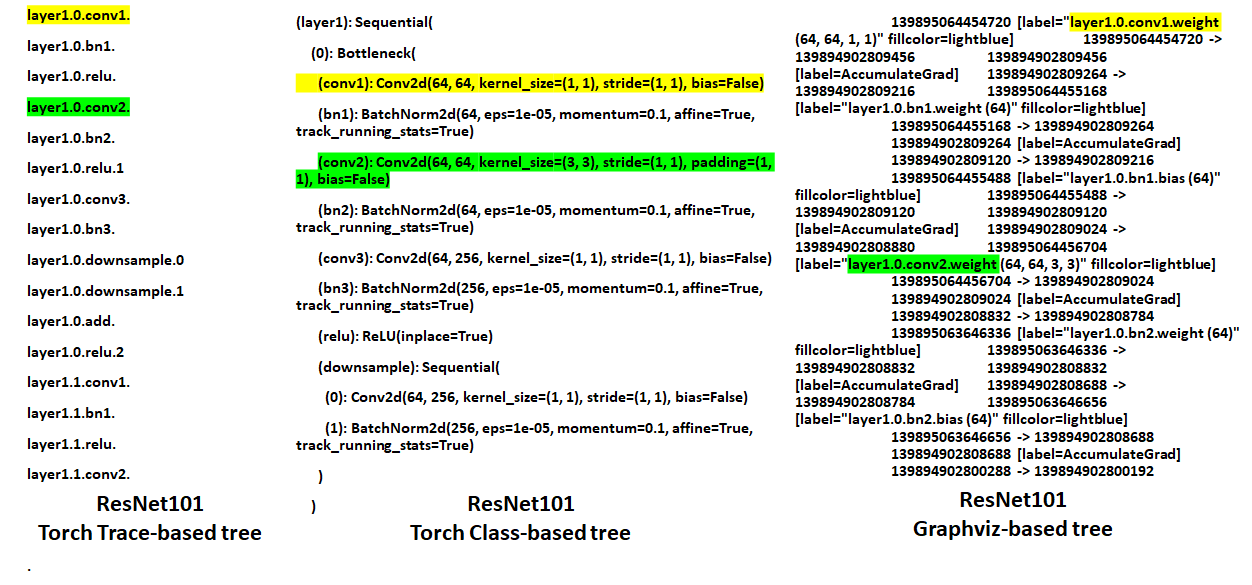}
  \centering
  \caption{ResNet101 graph representation (snippets) extracted by using (1) Torchvision library (left: trace-based tree, middle: block class-based tree) and (2) GraphViz library (right: nodes and edges). The highlighted text shows corresponding entities in the three tree representations.}
  \label{fig:05}
\end{figure}

To automate the utilization measurements, we identified computational units in all 35 image classification architectures supported by Torchvision \cite{torchvision2010}. The classification architectures of AI models are extracted into trace-based tree or block class-based tree textual representations - see Figure~\ref{fig:05} (left and middle). Using the TorchScript vision library \cite{torchvision2010}, these tree representations can be converted into a DiGraph representation \cite{digraph2019} (see Figure~\ref{fig:05} right) that can be pseudo-colored and visualized according to the block diagram shown in Figure~\ref{fig:04b}.

To simplify the automated placement of measurement probes (i.e., software hooks), we extracted a list of unique  computation units in block class-based trees of all Torchivision supported classification architectures and used the TensorState library \cite{tensorstate2020} for placing the measurement probes within each occurrence of a class. TensorState attaches measurement probes after each computation unit within a class and collects tensor-state statistics during image inferences. The placement of measurement probes after every computation unit yields a baseline for studying cost versus information tradeoffs.



\section{Variability of  Class Encodings in computation graph}
\label{variability}

Each trained AI model is unique in the way it encodes classes in computational units of computation graphs. We quantified the class encoding variability by averaging utilization values from three AI models trained on the same training dataset with varying random master seed for each training session and computing its standard deviation. The average utilization values varied between $0.5\,\%$ and $30\,\%$ with higher values in graph computation units before reaching layer3.  

\begin{figure}
\includegraphics[
  width=12cm,
  height=6cm,
  keepaspectratio,
]{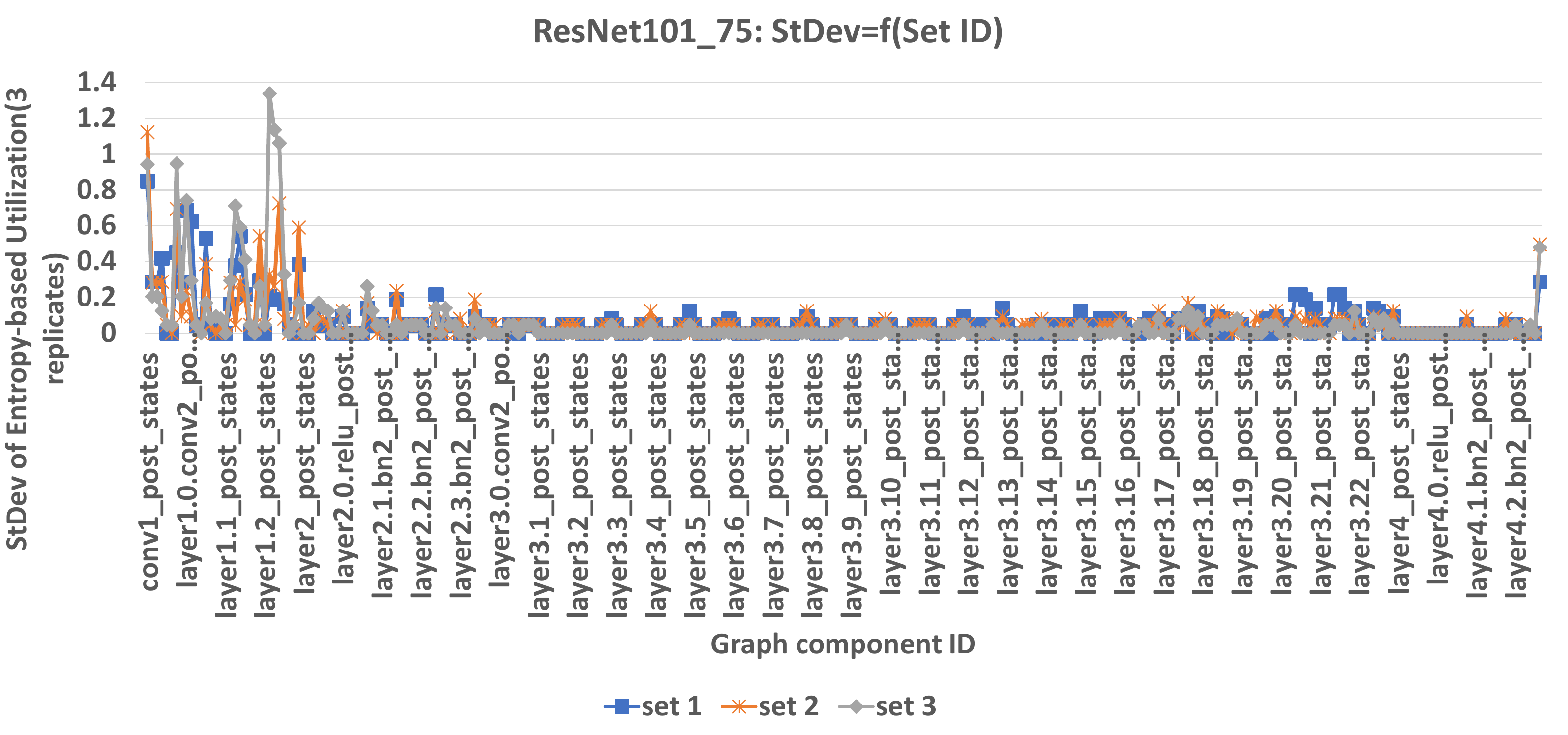}
  \centering
  \caption{Standard deviation of Entropy-based utilization at each graph location (probe) using the three sets of images defined in Table~\ref{table:03}.}
  \label{fig:10}
\end{figure}

Figure~\ref{fig:10} shows the standard deviation of utilization values per computational unit for three clean and three poisoned ResNet101 models predicting 75 traffic signs. The clean models were evaluated using clean images (Set 1) and resulted in three utilization values per graph component. The poisoned models were evaluated with clean images (Set 2) and poisoned images (Set 3) as defined in Table~\ref{table:03} and also resulted in three values per graph component per Set. The models were poisoned with Kelvin Instagram filter, which is boosting the earth tones of green, brown, and orange. The largest standard deviation was $1.34\,\%$ measured in layer1.2 by evaluating poisoned AI model on Set 3. The second largest value was $1.11\,\%$ measured in the first conv1  computation unit by evaluating poisoned AI model on Set 2. The larger magnitudes of utilization standard deviation in poisoned AI models over clean AI models suggest a higher variability in how trojans are encoded during each training session. 

\section{Reduction of Computational Requirements}
\label{computational_requirements}

\begin{figure}
\includegraphics[
  width=12cm,
  height=6cm,
  keepaspectratio,
]{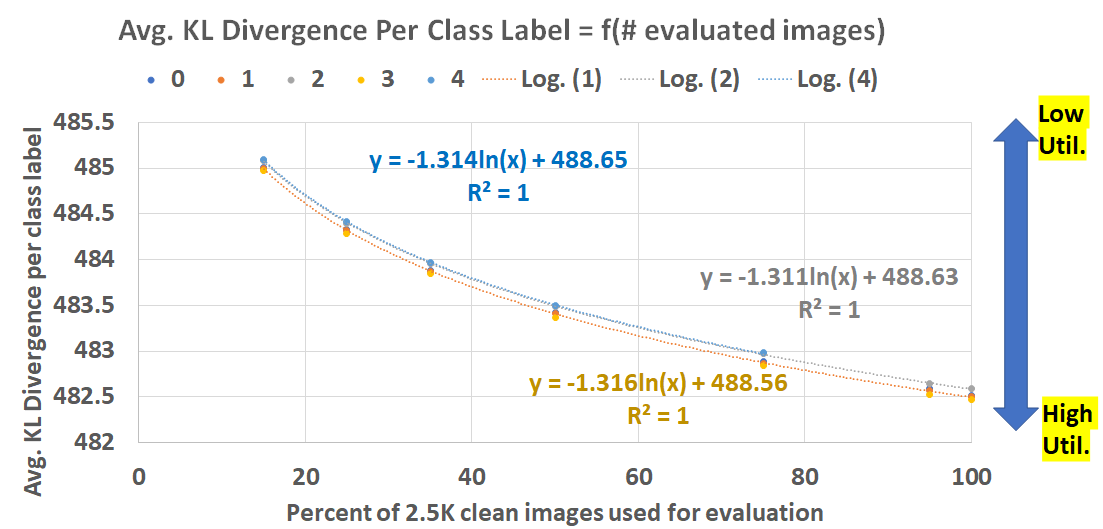}
  \centering
  \caption{Average KL Divergence utilization over all probes per class label as a function of reduced number of training images. The utilizations are evaluated on subsets between $15\,\%$ and $100\,\%$ of the 2.5K entire image collection for classes 0, 1, 2, 3, and 4. Three of the utilization curves are approximated with a ln function. 
}
  \label{fig:07}
\end{figure}

Following up on the computational complexity of utilization measurements in Section~\ref{method:complexity}, we gathered utilization values for varying number of images per class in order to build an extrapolation model and save some computational time. Figure~\ref{fig:07} shows average KL divergence-based utilization as a function of a reduced number of images per class. The extrapolation model is consistent across multiple predicted classes with average parameters   
$ \eta_{j}^{KL Div} = -1.314ln(M) + 488.61$
 where $M$ is the number of training images and the R-squared value equal to one ($R^{2} = 1$ implies high  trendline reliability).

The execution time  as a function of the number of evaluated images is approximated with a linear model: 
Time [s] $= 242.75*M - 497.02$ ($R^{2}=0.993$) where $M$ is the number of input images sampled at $\{15, 25, 35, 50, 95, 100\}$ percent of 2500 images per class. In our specific case, the time reduction was from 6.5 h for $100\,\%$ to 0.9 h for $15\,\%$ of evaluated images per class for a ResNet101 model predicting 40 traffic sign classes.

We also documented a computational tradeoff in ResNet101 between the number of measurement probes and the average computational time in Table~\ref{table:03z}. Based on the numerical values, the functional model is non-linear since the execution time depends not only on the number of probes but also on the complexity of the state tensors generated by each graph node.

\begin{table}
  \caption{Computation time of utilization as a function of the number of measurement probes  and placement in the ResNet101 architecture (sorted by time).}
  \label{table:03z}
\centering
\begin{tabular}{|c | c | c |}
\hline
\textbf{Probe name}	& \textbf{Number of probes} &	\textbf{Avg. Time [s]}   \\
\hline
\makecell{``layer3'' Block} & \makecell{1} & $80.67$  \\
\hline
\makecell{``conv2'' Node in all 4 ``layer'' Blocks} & 4 & $87.95$ \\
\hline
\makecell{Layer1.Bottleneck0 Block} & 1 & $88.03$ \\
\hline
\makecell{Fully Connected Block} &  \makecell{1} & $106.53$  \\
\hline
\makecell{Top ``conv1'' Node} & 1 & $107.65$  \\
\hline
\makecell{All Nodes ``ReLU''} & 34 & $258.95$  \\
\hline
\makecell{All in ``Conv2d'' Block} &   \makecell{104} & $280.26$           \\
\hline
\makecell{All Nodes} &   \makecell{286} & $	868.19$           \\
\hline
\end{tabular}
\end{table}

\end{document}